\documentclass[10pt,twocolumn,letterpaper]{article}

\usepackage{iccv}      %
\usepackage{times}
\usepackage{epsfig}

\usepackage[utf8]{inputenc}
\usepackage{graphicx,verbatimbox}
\usepackage{tcolorbox}
\usepackage{rotating}

\usepackage{tikz}
\usepackage{comment}
\usepackage{amsmath,amssymb} %

\usepackage{booktabs} %
\usepackage{xcolor}
\usepackage{color,colortbl}
\usepackage{pifont}
\usepackage[pagebackref,breaklinks,colorlinks]{hyperref}

\usepackage{multirow}
\usepackage[font=footnotesize]{caption}
\usepackage{placeins}
\usepackage{enumitem}%

\iccvfinalcopy 

\usepackage{xspace}

\definecolor{Goldenrod}{RGB}{245,245,220}

\newcommand{\cmarkgr}{\textcolor{teal}{\ding{51}}\xspace}%
\newcommand{\xmarkgr}{\textcolor{purple}{\ding{55}}\xspace}%

\def\mypar#1{\vspace{0.4em}{\noindent\bf #1.}\hspace{1mm}}

\newcommand{\diff}[1]{{\footnotesize{#1}}}

\newcommand{\ours}{{cosub}\xspace}

\newcommand{\SD}{\tau}

\def \pzo {\phantom{0}} 
\def \dzo {\phantom{00}} 
\def \tzo {\phantom{000}}

\def \T  {{\mathcal T}}

\def\confYear{2023}

\title{Co-training $2^L$ Submodels for Visual Recognition}
\makeatletter
\let\inserttitle\@title
\makeatother

\author{
\begin{minipage}{\linewidth}
\centering
\normalsize Hugo Touvron$^{\star,\dagger}$ \hspace{0.25cm}  Matthieu Cord$^{\dagger}$ \hspace{0.25cm}  Maxime Oquab$^{\star}$ \hspace{0.25cm}  Piotr Bojanowski$^{\star}$ \hspace{0.25cm} Jakob Verbeek$^{\star}$ \hspace{0.25cm}  Herv\'e J\'egou$^{\star}$ \\[0.3cm] 
\scalebox{1.}{$^\star$Meta AI / FAIR Paris\hspace{1.6cm} $^\dagger$Sorbonne University}\\
\end{minipage}
}

\begin{document}

\maketitle

\begin{abstract} 
We introduce submodel co-training, a regularization method related to co-training, self-distillation and stochastic depth. 
Given a neural network to be trained, for each sample we implicitly instantiate two altered networks, ``submodels'', with stochastic depth:
we activate only a subset of the layers. 
Each network serves as a soft teacher to the other, by providing a loss that complements the regular loss provided by the one-hot label. 
Our approach, dubbed ``\ours'', uses a single set of weights, and does not involve a pre-trained external model or temporal averaging. 

Experimentally, we show that submodel co-training is effective to train backbones for recognition tasks such as image classification and semantic segmentation. Our approach is compatible with multiple architectures, including RegNet, ViT, PiT, XCiT, Swin and ConvNext.  
Our training strategy improves their results in comparable settings. 
For instance, a ViT-B pretrained with \ours on ImageNet-21k obtains 87.4\% top-1 acc.\ @448 on ImageNet-val. 
\end{abstract}

\vspace{-1em}

\section{Introduction}

\begin{figure}[t]
\vspace{-0.77em}
\centering
\includegraphics[width=.59\linewidth]{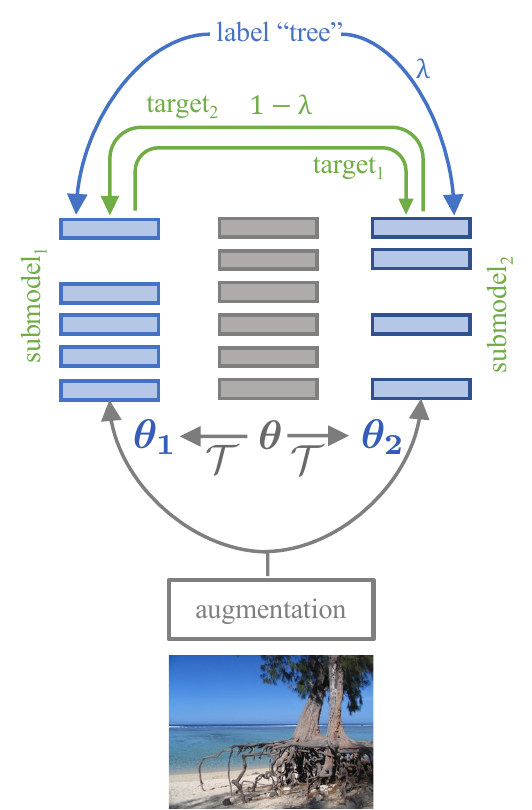}
\caption{
\textbf{Co-training of submodels (cosub):} for each image, two submodels are sampled  by randomly dropping layers of the full model. 
The training signal for each submodel mixes the  cross-entropy loss from the image label with a self-distillation loss obtained from the other submodel.
\label{fig:splash}}
\end{figure}

Although the fundamental ideas of deep trainable neural networks have been around for decades, only recently have barriers been removed to allow breakthroughs in successfully training deep neural architectures in practice.
Many of these barriers are related to non-convex optimization in one way or another, which is central to the success of modern neural networks. 
The optimization challenges have been addressed from multiple angles in the literature. 
First, modern architectures are designed to facilitate the optimization of very deep networks. 
An exceptionally successful design principle is using residual connections~\cite{He2016ResNet,He2016IdentityMappings}. 
Although this does not change the expressiveness of the functions that the network can implement, the improved gradient flow alleviates, to some extent, the difficulties of optimizing very deep networks. 
Another key element to the optimization is the importance of data, revealed by the step-change in visual recognition performance resulting from the ImageNet dataset~\cite{deng2009imagenet}, and the popularization of transfer learning with pre-training on large datasets~\cite{oquab2014learning,Yosinski2014HowTA}. 

However, even when (pre-)trained with millions of images,  recent deep networks with  millions if not  billions of parameters, are still heavily overparameterized. 
Traditional regularization like weight decay, dropout~\cite{srivastava2014dropout}, or label smoothing~\cite{Szegedy2016RethinkingTI} are limited in their ability to address this issue.  
Data-augmentation strategies, including those mixing different images like Mixup~\cite{Zhang2017Mixup} and CutMix~\cite{Yun2019CutMix}, have proven to provide a complementary data-driven form of regularization.
More recently, multiple works propose to resort to self-supervised pre-training. 
These approaches rely on a proxy objective that generally provides more  supervision signal than the one available from labels. 
For instance, recently there has been renewed interest in (masked) auto-encoders~\cite{bao2021beit,He2021MaskedAA,el2021training}, which were popular in the early deep learning literature~\cite{bourlard88bc,erhan2010does,hinton93nips}. 
Similarly, contrastive approaches \cite{he2019moco,caron2021emerging}
provide a richer supervision less prone to a supervision collapse~\cite{doersch2020crosstransformers}. 
Overall, self-supervised learning makes it possible to learn larger models with less data, possibly reducing the need of a pre-training stage~\cite{el2021large}. 
 
Distillation is a complementary approach
to improve optimization.
Distillation techniques were originally developed to transfer knowledge from a teacher model to a student model~\cite{ba14nips,Hinton2015DistillingTK}, allowing the student to improve over learning from the data directly.
In contrast to traditional distillation, co-distillation does not require pre-training a  (strong) teacher. 
Instead, a  pool of models supervise each other. 
Practically, it faces several limitations, including the  difficulty of jointly training more than two students for complexity reasons, as it involves duplicating the weights. 

In this paper, we propose a practical way to enable co-training for a very large number of students. 
We consider a single target model to be trained, and we instantiate two submodels \emph{on-the-fly}, simply by layerwise dropout~\cite{Huang2016DeepNW,fan2019reducing}. 
This gives us two  neural networks through which we can backpropagate to the shared parameters of the target model.
In addition to the regular training loss, each submodel serves as a teacher to the other, which provides an additional supervision signal ensuring the consistency across the submodels. Our approach is illustrated in Figure~\ref{fig:splash}: the parameter $\lambda$ controls the importance of the co-training loss compared to the label loss, and our experiments show that it significantly increases the final model accuracy. 

This co-training across different submodels, which we refer to as \emph{\ours}, can be regarded as a massive co-training between $2^L$ models that share a common set of parameters, where $L$ is the number of layers in the target architecture. 
The target model can be interpreted as  the expectation of all models. 
With a layer drop-rate set to 0.5, for instance for a ViT-H model, all submodels are equiprobable, and then it amounts to averaging the weights of $2^{2\times32}$ models.

\noindent 
Our contributions can be summarized as follows:
\begin{itemize}[noitemsep,topsep=2pt]
\item We introduce a novel training approach for deep neural networks: We co-train submodels. This significantly improves the training of most models, establishing the new state of the art in multiple cases. For instance, after pre-training ViT-B on Imagenet-21k and fine-tuning it at resolution 448, we obtain 87.4\% top-1 accuracy on Imagenet-val.
\item We provide an efficient implementation to subsample models on the fly. It is a simple yet effective variation of stochastic depth \cite{Huang2016DeepNW} to drop residual blocks.
\item We provide multiple analyses and ablations. Noticeably, we show that our submodels are effective models by themselves even with significant trimming, similar to LayerDrop~\cite{fan2019reducing} in natural language processing.
\item We validate our approach on multiple architectures (like ViT, ResNet, RegNet, PiT, XCiT, Swin, ConvNext), both for image classification --trained from scratch or with transfer--, and semantic segmentation. 
\item We will share models/code for reproducibility in the \href{https://github.com/facebookresearch/deit}{\textcolor{black!60}{DeiT repository}}. 
\end{itemize}

\section{Related work}

\begin{figure*}
    \centering
    \includegraphics[width=\linewidth]{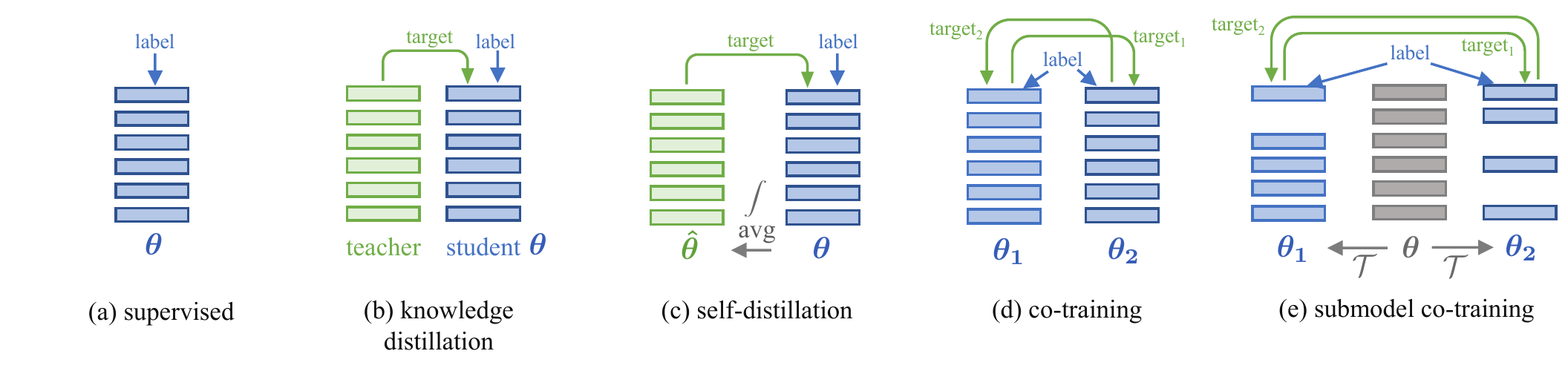}
    
    \vspace{-0.5em}
    \begin{minipage}{0.63\linewidth}
    \caption{
    Brief summary of related works and of submodel co-training (\ours). (a) supervised baseline: the supervision is provided as a one-hot label. (b) Knowledge distillation~\cite{Hinton2015DistillingTK}: a teacher provides a (soft) target. A simple yet effective variant~\cite{Touvron2020TrainingDI}  combines the predicted label of the teacher with the image label. KD requires a pre-existing teacher that must be stored for inference. (c) In self-distillation, the teacher is obtained from the model itself, typically by model averaging. (d) Co-training involves two distinct models that serves as teacher to each other. (e) Submodel co-training generates on-the-fly two distinct models (amongst $2^L$) for each sample, which serves a teacher to each other. We only need to store a single set of weights for the model and optimizer because the random selections $\theta_1$\,=\,$\mathcal{T}(\theta)$ (resp., $\theta_2$) of submodels allow us to back-propagate on~$\theta$. 
    \label{fig:codist}}    
    \end{minipage}
    \hfill
    \raisebox{1.4em}{
    \begin{minipage}{0.35\linewidth}
    {\footnotesize
    \begin{tabular}{@{\ }lc@{\quad\ }cc@{\ }}
                       & \multicolumn{2}{c}{model params} \ \ ~    &   \\
    method             & weights      & optimizer  &  compute$^\star$ \\
    \midrule 
    (a) supervised     & $\times 1$ &   $\times 1$  &  $\times 1$ \\
    (b) KD$^{\dagger}$  & $\times 2$ &   $\times 1$  &  $\times 1.33$ \\
    (c) mean teacher   & $\times 2$ &   $\times 1$  &  $\times 1.33$ \\
    (d) co-training    & $\times 2$ &   $\times 2$  &  $\times 2$ \\
    (e) Cosub (ours)   & $\times 1$ &   $\times 1$  &  $\times 2$ \\
    \bottomrule
    \end{tabular}
    
    \makebox{$^\star$backward pass assumed 2$\times$ the complexity of forward} \\[-3pt]
    $^{\dagger}$requires a trained teacher model}
    \end{minipage}
    }
    \vspace{-1.5em}
\end{figure*}

\mypar{Knowledge distillation}  
Originally, distillation was  introduced as a way to train a model such that it reproduces the performance of another model~\cite{ba14nips,Hinton2015DistillingTK}.
The typical use-case is to improve the 
quality of a %
relatively small model by leveraging a strong teacher, whose complexity may be prohibitive for a practical deployment. 
The teacher's soft labels have a similar effect as label smoothing~\cite{Yuan2019RevisitKD}. 
As shown by Wei \etal~\cite{Wei2020CircumventingOO}, the teacher's supervision takes into account the effects of the data augmentation, which sometimes causes a misalignment between the real label and the image. 
Knowledge distillation can transfer inductive biases~\cite{abnar2020transferring}  in a soft way in a student model when  using a teacher model these biases are enforced in a hard way. Touvron~\etal~\cite{Touvron2020TrainingDI} proposed a variant of distillation adapted to Vision Transformer (ViT), showing the effectiveness of using a ConvNet teacher for a ViT student. 

\mypar{Mixture models} 
Ensembling has a long history that we can trace back to the origins of statistics~\cite{McLachlan00} and the possibility of improving the precision of measurements with multiple observations. 
In machine learning, bagging~\cite{breiman1996bagging} combines multiple weak classifiers to produce a strong one. This idea is naturally extended to neural networks, where it can offer improved stability or accuracy, or other properties like anytime inference~\cite{ruiz2021anytime}. 
A mixture  model can also be seen as a larger model with an internal parallel structure. 

\mypar{Co-distillation} 
At the interface of mixture models and distillation, the concept of co-distillation \cite{zhu2018knowledge,zhang2018deep} does not require a prior teacher: the mixture serves as the teacher to all the networks in the mixture. 
The models are jointly optimized, which leads to improved  accuracy of the individual models~\cite{anil2018large}. 
This can be regarded as a form of collaborative learning between the different elements of the mixture~\cite{song2018collaborative}.
Compared to training only one model, co-distillation involves training two or more models, each requiring storage for weights and activation, and computing backward and forward passes.

\mypar{Exponential moving average teacher} 
A special form of teacher is a model obtained from the student model itself. Tarvainen and Valpila~\cite{tarvainen2017mean} show that such a model obtained by temporal averaging of intermediate checkpoints during the training is an effective teacher, that  can be obtained with an exponential moving average (EMA). This idea has also been adopted in self-supervised training with learning schemes  like  DINO~\cite{caron2021emerging}. %
This method involves storing a copy of the weights corresponding to the temporal averaging carried out by the EMA model. %

\mypar{Dropout and  model populations}
Dropout~\cite{srivastava2014dropout} is an effective way to regularize models. With residual architectures, one effective way to train deeper architecture is stochastic depth~\cite{Huang2016DeepNW}. It  reduces the size of a network at training time by dropping  residual blocks during training. 
The initial goal of this approach was to improve the training of deep network. However, there are other implications: in natural language processing, Fan \etal show that the counterpart of stochastic depth, namely LayerDrop~\cite{fan2019reducing}, is effective to reduce the transformer depth on demand. Interestingly, submodels extracted from a model trained with LayerDrop are stronger than those trained with the full target depth with a proper selection strategy for layers.  

In our paper, we regard stochastic depth as a regularization technique, but at the same time we adopt the point of view of LayerDrop, \ie, an effective way to train a population of submodels that share parameters, or  entire layers in our case. 
From that viewpoint, stochastic depth amounts to training $2^L$ distinct models.  
This is different from population training as involved in network space design~\cite{radosavovic2020designing}, whose goal is  related to network architecture search~\cite{elsken2019neural}, which aims at optimizing the architecture itself.  

More recently, model soups \cite{wortsman2022model} are models obtained by averaging the weights of multiple models finetuned with different hyperparameter configurations. The authors show that it often improves accuracy and robustness. 
We point out that stochastic depth can be regarded as a special form of model soup over the entire population of submodels that we can instantiate with stochastic depth. We further discuss this relationship in the next section. 
\section{Submodel co-training} 
\label{sec:codist}

In this section, we present our \emph{\ours} approach, namely co-training submodels.   
We consider a neural network $f_{\theta}$ parameterized by learnable parameters $\theta$. 

\mypar{Submodel instantiation: model augmentation operator} 
We first define a model augmentation operator $\T$. For a given neural network, it provides a set of parameters $\theta'=\T(\theta,R)$ that allow us to define variations $f_{\theta'}$ of the function $f_\theta$ by drawing a random variable $R$. 
The model augmentation is such that $f_{\theta}$ and $f_{\theta'}$ share parameters, hence any update on $\theta'$ modifies $\theta$ and therefore $f_\theta$ accordingly. A simple way to define $\T$ is to replace some parameters by zeros, which corresponds to dropout~\cite{srivastava2014dropout}. In this paper, we focus on stochastic depth, which has interesting connections with model averaging~\cite{wortsman2022model}.

\mypar{Overview} 
For a given  training sample $x$ within a batch, the training is as follows (see also Figure~\ref{fig:splash}): 
\begin{enumerate}[noitemsep,topsep=2pt]
    \item We first data-augment the image, producing $\hat{x}$. 
    \item The image is duplicated with the batch, effectively doubling the batch size, which hence contains two identical copies of each augmented image. 
    \item We determine the stochastic depth pattern for each sample according to the target drop-rate $\SD$, which amounts to producing two functions $f_{\theta_1}$ and $f_{\theta_2}$. See Section~\ref{sec:esd} for the details of this procedure. 
    \item The forward pass proceeds as usual: we compute the soft output labels $y_1=f_{\theta_1}(\hat{x})$ and $y_2=f_{\theta_2}(\hat{x})$. 
    \item We compute the losses and the  backwards pass accordingly. Note that the gradients  on $\theta_1$ and $\theta_2$ are  directly used to update $\theta$, as they are just subsets of $\theta$.
\end{enumerate}

\mypar{Loss} 
Each submodel is trained using a weighted average of (i) the standard binary cross-entropy loss obtained from the image label $y$, and (ii) a binary cross-entropy loss \wrt the soft-labels computed for the the same image  by the other submodel.
The respective weight of the standard binary cross-entropy loss ${\mathcal L}_\mathrm{label}$ versus the cosub loss ${\mathcal L}_\mathrm{cosub}$  is controlled by the hyper-parameter $\lambda$, as 
\begin{align}
{\mathcal L}_\mathrm{tot} & = \lambda \, {\mathcal L}_\mathrm{label} + (1-\lambda)\,{\mathcal L}_\mathrm{cosub}. 
\end{align}
In details, the loss writes as 
\begin{align}
    {\mathcal L}_\mathrm{tot} & = \lambda \left( \frac{{\mathcal L}(f_{\theta_1}(y_1,y)) +  {\mathcal L}  (f_{\theta_2}(y_2,y))}{2} \right)\nonumber\\
                  & + (1-\lambda) \left(\frac{{\mathcal L}(y_1,\mathrm{s_g}(y_2)) + {\mathcal L}(y_2,\mathrm{s_g}(y_1))}{2}  \right), 
\end{align} 
where ${\mathcal L}(y,y')$ is either the binary cross-entropy (BCE) or a cross-entropy (CE) loss. Importantly, note that when applying this loss, we do not back-propagate on the second term $y'$, which is represented by the stop-gradient operator $\mathrm{s_g}(\cdot)$. 

\mypar{Discussion} 
The submodel instantiation provides a model that is, by itself, a valid neural network model. 
Co-training submodels is a way to enforce that all such submodels produce a consistent output. 
Amongst these submodels, a very special case is when we retain all residual blocks, which is the model that we primarily intend to  use at inference time. 
Note that  Fan \etal~\cite{fan2019reducing} show, in an NLP context, that submodels extracted from a deeper model are superior, for a given target depth.
With proper rescaling of the residual branches, this specific submodel with all blocks activated can be seen as the expectation over all models. 
This can be shown as follows: let us consider an extra scalar parameter $s_l$ associated with a given residual block $r_l(x)$, which we use as a multiplicative factor on output of each residual: $s_l=1$ if the residual block $r_l$ is included in the submodel, $s_l=0$ otherwise. 
Each submodel is fully parameterized by a binary vector $s=(s_1,\dots,s_L)$, where $L$ is the total number of layers. Therefore all submodels have the same parameters $\theta$, and only differ by $s$ indicating the zeroed residual blocks. The weight expectation $\bar{\theta}$ is hence
\begin{align}
    \bar{\theta} & ={\mathbb E}_{s\in [0,1]^L}[\theta] = \theta, \nonumber \\
    \bar{s}  & =  {\mathbb E}_{s\in [0,1]^L}[s] = [1-\tau,\dots,1-\tau]^\top, 
\end{align}
where $\bar{s}$ is the scaling factor used in stochastic depth. Under a uniform distribution of submodels that is obtained when the stochastic drop rate $\tau=0.5$, the inference-time model is exactly the average of all $2^L$ models. While stochastic depth is often regarded as a regularization technique, this averaging interpretation relates it to the recent model soup~\cite{wortsman2022model}.

\section{Efficient stochastic depth}
\label{sec:esd}

We  revisit the original stochastic depth~\cite{Huang2016DeepNW} formulation in order enable an efficient implementation, which we will release for PyTorch~\cite{paszke2019pytorch}.
Instantiating a submodel for a sample amounts to selecting a subset of residual layers of the model and performing training on these. This can be implemented using the stochastic depth (SD) approach, whose objective was initially to improve the training of very deep networks. 
In stochastic depth, for each sample and each layers of the network, we select whether the layers will be dropped or not with a certain probability $\SD$.

In practice, \eg in the timm library~\cite{pytorchmodels}, stochastic depth is implemented by masking with zeros the residuals added for a given sample of a batch. 
This is not efficient: this naive approach performs the computations for a residual then throws it away, wasting computation. 

\begin{figure}
    \begin{minipage}{\linewidth}

    \includegraphics[width=\linewidth]{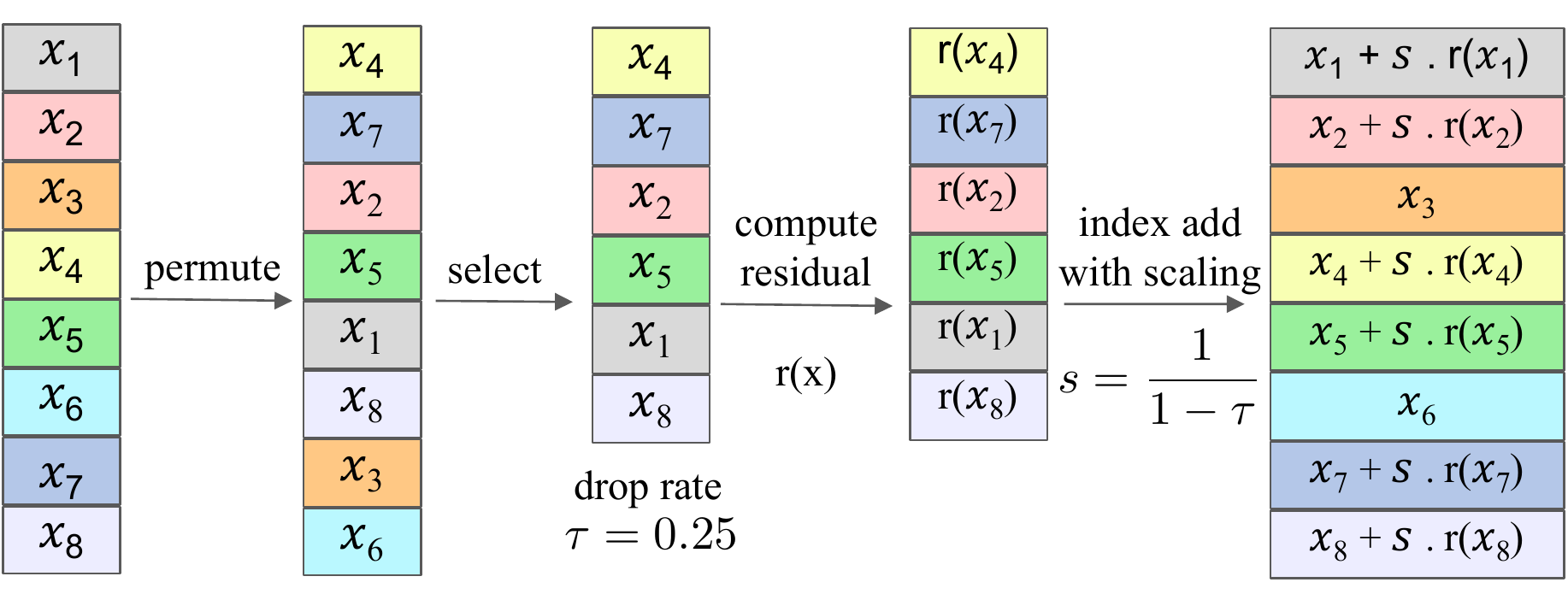}
    \end{minipage}
        \vspace{-0.5em}
    \caption{Efficient implementation of stochastic depth, using the permute-select approach. In this example, we drop the residuals for samples 3 and 6, corresponding to a drop rate $\SD$\,=\,0.25. For a drop rate higher than 0.1, the overhead of the approach is negligible in terms of  memory and compute. 
        \label{fig:efficient_sd}}
\vspace{-0.7em}
\end{figure}

Our efficient stochastic depth (ESD) approach addresses this issue, saving both memory and compute, and is illustrated in Figure~\ref{fig:efficient_sd}.
For each layer, given a batch size $B$ and a drop rate $\SD$, we apply the layer to  $B_\mathrm{keep} = \text{round}(B \times (1 - \SD))$ samples in the batch. In contrast to the original version of SD, our efficient version drops a fixed number of samples at each layer, where $d$ is adjusted such that $B_\mathrm{keep}$ is an integer. In our experiments, this did not have an effect on the final accuracy of the models.
Our efficient implementation proceeds as follows: (i) we apply a random permutation of the $B$ samples, then (ii) we select the first $B_\mathrm{keep}$ samples. We then compute the residual function for the selected subset, then add the result onto the full batch using the built-in \texttt{index\_add} function and scaling the result by $\frac{1}{1-\SD}$ %
to have the correct scaling as  in the original SD formulation.

\mypar{Discussion: progressive vs uniform rate} 
In the original paper~\cite{Huang2016DeepNW}, stochastic depth drops a layer with a probability that is linearly increasing during the forward pass: layers closer to the output has a higher chance to be dropped. However this strategy is limited with high drop rates, as later layers are excessively dropped. 
Touvron \etal~\cite{touvron2021going} adopt a uniform rate drop per layer. It is as effective as the progressive rate with vision transformers, but it makes it possible to target a drop-rate greater than 0.5 on average. For this reason we adopt an uniform drop rate everywhere. 

Our efficient stochastic depth implementation variant works both with uniform and progressive drop-rate. In the case of progressive, the effective batch size is decreasing during to the forward pass. One limitation is that our technique implies a %
quantization of the drop-rate, which can be problematic for small batch sizes. Therefore one must take care to properly verify that the drop-rate is not too severely quantized, in particular in the progressive case for which the drop-rate can be very high in the last layers.
In the supplemental material (Appendix~\ref{app:esd}), we discuss with more details the drop-rate quantization effect.

\section{Experiments}

We evaluate our approach on residual architectures~\cite{He2016ResNet,He2016IdentityMappings}, since they are readily compatible with the stochastic dropout.  We take as our main reference the vanilla Vision Transformer introduced by Dosovitskiy \etal~\cite{dosovitskiy2020image}. 
However, as shown  in our experiments, our approach is effective with all the residual architectures that we have considered.

\subsection{Baseline and training settings}

We adopt the state-of-the-art training procedure proposed in the DeiT III paper~\cite{touvron2022deitIII} as baseline for transformers architectures and the training procedure from Wightman \etal~\cite{wightman2021resnet} for the convnets. 
All hyper-parameters are identical except on Imagenet-21k, where the hyper-parameter $\tau$ are adjusted depending on the training setting. We recapitulate the training hyper-parameters in Appendix~\ref{app:training_details}. 

 We additionally adopt and measure the impact of LayerDecay for the fine-tuning when transferring from Imagenet-21k to Imagenet-1k. This method slightly boosts the performance, as discussed later in this section, and in Table~\ref{tab:ablation_21k}. This method was adopted in multiple recent papers and in particular for fine-tuning of self-supervised approaches~\cite{bao2021beit,He2021MaskedAA}, yet the contribution of this fine-tuning ingredient was not quantitatively measured.

\begin{figure}
    \begin{minipage}{0.48 \linewidth}
    ~ \hfill \quad \small ViT-B ($\SD=0.2$) \hfill  ~
    
    \includegraphics[height=\linewidth]{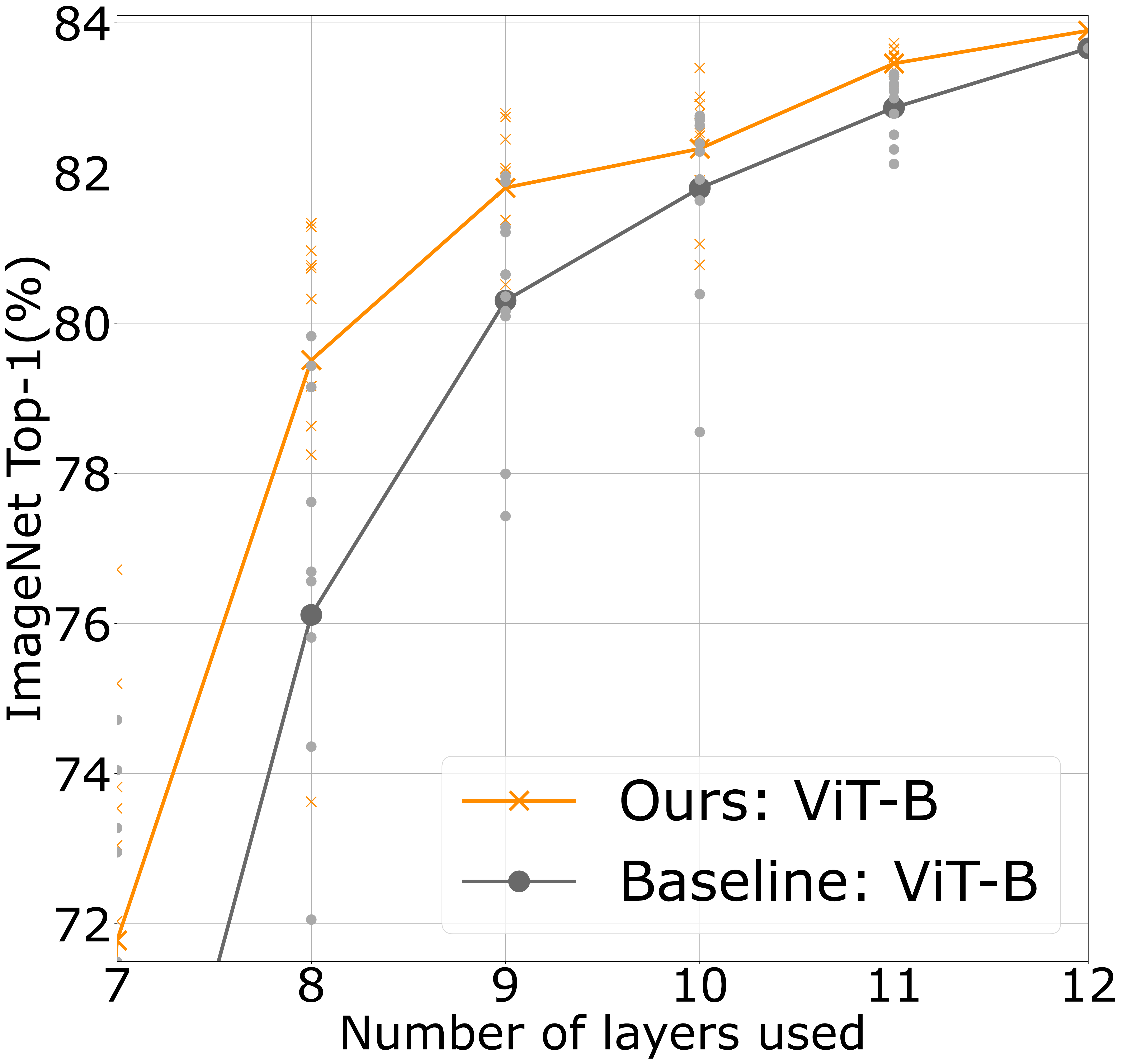}
    \end{minipage}
    \hfill
    \begin{minipage}{0.48 \linewidth}
    ~ \hfill \small ViT-L ($\SD=0.45$) \hfill  ~
    
    \includegraphics[height=\linewidth,trim={70pt 0 0 0},clip]{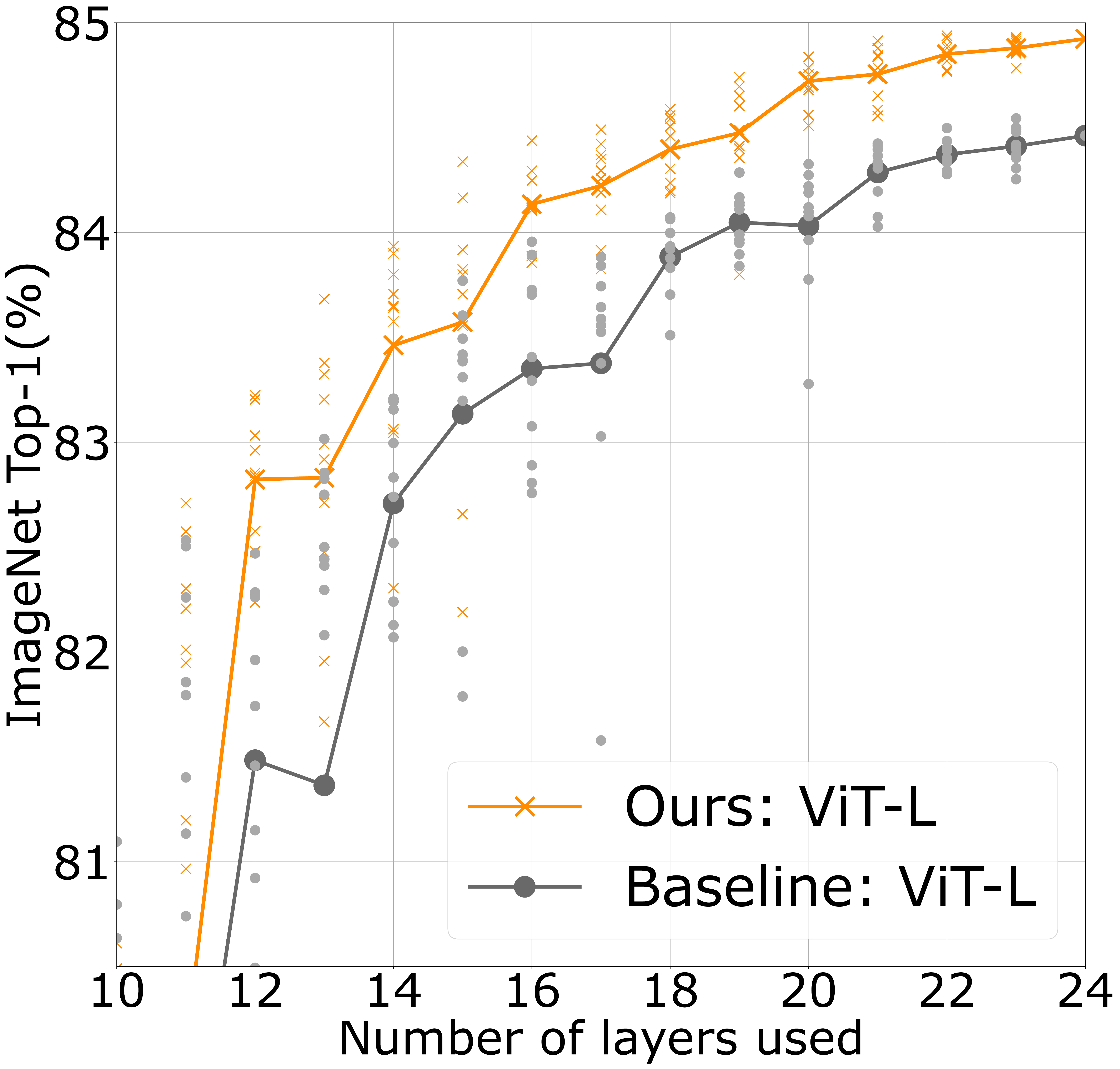}
    \end{minipage}
    \vspace{-0.5em}
    \caption{Cosub as population training: each submodel extracted by a transformation $\T$ is a valid neural network. Our \ours strategy can hence be regarded as co-training a large number of subnetworks. We plot the accuracy of the submodels as a function of the number of layers that we preserve. 
    We drop layers with probability $\tau$\,=\,$0.2$ for ViT-B and $\tau$\,=\,$0.45$ for ViT-L.  
    On average our method provides a significant boost in performance for the whole population of submodels extracted from the main model. 
        \label{fig:sub_mod}}
\end{figure}

\begin{table}[t]
    \centering
    \scalebox{0.8}{
    \begin{tabular}{lccccc}
        \toprule
          Model    &  \ \ ViT-S\ \  & \ ViT-M\  & \ ViT-B\  & \ \ ViT-L\ \  & \ ViT-H \   \\
          $\SD$  & ($0.05$) &  ($0.1$) & ($0.2$) & ($0.45$)& ($0.6$) \\
         baseline & 81.4 & 82.5 & 83.7 & 84.5 & 84.9  \\
         \rowcolor{blue!7}
         baseline + \ours \ \ & 81.5 & 82.8 &  83.9 & 84.9 & 85.5 \\
         \bottomrule
    \end{tabular}}
    \caption{Comparison of top-1 accuracy with ViT architecture trained with/without \ours~ at resolution $224$ (800 epochs) on Imagenet-1k. 
    The improvement is not significant for the smaller architecture ViT-S. Cosub is gradually more effective when we increase the model size and the SD rate.  
    \label{tab:model_size}}
\end{table}

\begin{table}
    \centering
    \scalebox{0.8}{
    \begin{tabular}{lccccc}
    \toprule
         Model & \ $\SD$ \  & Original & Baseline & \ +\ours\  & \ $\Delta$ \  \\
         \midrule
         \multicolumn{6}{c}{\bf Transformers}\\[1pt]
         ViT-L~\cite{dosovitskiy2020image}  & 0.45 & 76.5 & 84.5 & \cellcolor{blue!7} 84.9 & \diff{+0.4} \\
         CaiT-L24~\cite{touvron2021going} & 0.45 & \_   & 83.8 &  \cellcolor{blue!7} 84.4 & \diff{+0.6} \\
         PiT-B~\cite{Heo2021RethinkingSD}  & 0.25 & 82.0   & 83.8 &  \cellcolor{blue!7} 84.1 & \diff{+0.3} \\
         XCiT-L12~\cite{el2021xcit}  & 0.20 &  \_  & 82.6 &  \cellcolor{blue!7} 83.0 & \diff{+0.4} \\
         Swin-B~\cite{liu2021swin}  & 0.20 & 83.5  & 82.9 &  \cellcolor{blue!7} 83.3 & \diff{+0.4}\\
         Swin-L~\cite{liu2021swin}  & 0.45 & \_   & 80.8 &  \cellcolor{blue!7} 84.0 & \diff{+3.2}\\
              \\[-0.9em]
         \multicolumn{6}{c}{\bf Convnets}\\[1pt]
         ResNet-50~\cite{He2016ResNet} & 0.10 & 76.2 & 80.2 & \cellcolor{blue!7} 80.3 & \diff{+0.1}\\
         ResNet-101~\cite{He2016ResNet} & 0.20 & 77.4 & 81.8 & \cellcolor{blue!7} 82.1 & \diff{+0.4}\\
         ResNet-152~\cite{He2016ResNet} & 0.30 & 78.3 & 82.4 & \cellcolor{blue!7} 83.1 & \diff{+0.7}\\
         RegNet-16GF~\cite{Radosavovic2020RegNet} \quad\,  & 0.30 & 80.4 & 82.9 & \cellcolor{blue!7} 83.8 & \diff{+0.9}\\
\bottomrule

    \end{tabular}}
    \vspace{-0.7em}
    \caption{Benefit of \ours for different architectures trained from scratch on Imagenet-1k at resolution $224$. We  report top-1 acc. for the supervised baseline  and cosub, as well as  results reported in the corresponding papers  when available (trained with different settings). 
    We have adjusted the stochastic-depth drop-rate (SD) hyper-parameter for each architecture. 
    \label{tab:comp_archi}}
\end{table}

\begin{table}
    \centering
    \scalebox{0.8}{
    \begin{tabular}{rccc}
    \toprule
         Loss & \cellcolor{blue!7} \  BCE-soft \ & \ BCE-hard \ & \ CE-hard \ \\
         \midrule
         Imagenet-val top-1 accuracy  & \cellcolor{blue!7} 83.5 & 83.5 &  81.9 \\
         \bottomrule
    \end{tabular}
    }
    \vspace{-0.7em}
    \caption{Ablation on the loss for \ours with ViT-H trained at resolution $126\!\times\!126$ on Imagenet-1k during 800 epochs. The training is inherited from DeiT-III, which also uses BCE when training with Imagenet-1k only. }
    \label{tab:loss_abl}
\end{table}

\subsection{Empirical analysis of \ours}

We perform various ablations on Imagenet-1k to analyse the impact of our training method on the learned networks.

\mypar{Performance for different model sizes and architectures} 
Table~\ref{tab:model_size} provides the results obtained by the baseline and \ours when we vary the model size of vision transformers. The stochastic depth coefficient was optimized for the baseline and we keep it unchanged with \ours. As to be expected, our method is almost neutral for small models like ViT-S: +0.1\% top-1 accuracy, which is about the standard deviation of measurements. The improvement is increasingly important for larger models, up to a significant improvement of +0.6\% top-1 accuracy for ViT-H models. 

In Table~\ref{tab:comp_archi}, we show that our approach is beneficial with all architectures that we have tried. We report the results of the original paper, evaluate the performance with our baseline training, and measure the improvement brought by \ours. For almost all architectures, we observe a significant boost in performance. The exception is the ResNet-50, for which \ours improves the top-1 accuracy by only +0.1\%, similar to our observation with ViT-S. 
In Table \ref{tab:comp_archi_imnet21k} in the appendix  we present improved results obtained for multiple architectures pre-trained with \ours on Imagenet-21k.

\mypar{Analysis of submodel performance}
With \ours, we sample different subnetworks during  training to  performed the co-training. 
We analyse the impact of \ours on the accuracy of the sub networks themselves. 
In Figure~\ref{fig:sub_mod} we consider the accuracy of submodels of different size of ViT-B and ViT-L models.
Cosub improves the accuracy of the whole population of sub-networks and, in particular, the target network.

\mypar{Loss formulation}
In Table~\ref{tab:loss_abl} we experiment with  different losses for \ours. 
With Imagenet1k, DeiT-III training uses BCE instead of CE for the main loss. 
With \ours, BCE is more compatible with the loss of the baseline and, as to be expected, we also observe a better performance with BCE.
We have done ablations using hard and soft targets for the \ours loss.
The results are similar, therefore by default we keep soft-targets for the \ours loss.

\begin{table}
    \centering
    \scalebox{0.8}{
    \begin{tabular}{clcc}
        \toprule
         & Method &  \quad   ViT-L  \quad \  & \ \ \   ViT-H  \ \ \ \\
         \midrule
          (a) & supervised baseline~\cite{touvron2022deitIII} & 84.5 & 84.9 \\
          (b) & KD           & 85.3 & 85.3 \\
          (c) & mean teacher & 84.4 & 83.4 \\
          (d) & co-training  & 82.6 & 83.1 \\
          \rowcolor{blue!7}
          (e) & \ours        & 84.9 & 85.5 \\
          \quad  (b) + (e) \quad \  & KD + \ours & 85.3 & 85.7 \\
          \bottomrule
          
    \end{tabular}}
    \vspace{-0.7em}
    \caption{Training strategies with distillation. We compare on Imagenet-1k at resolution $224\times224$ different  approaches involving co- or self-training with distillation. 
    KD, mean teacher and co-training use the same $\lambda=0.5$ and same hyper-parameters as in Deit-III. Unlike \ours, the mean teacher (c) requires other hyper-parameters for EMA. We perform a small grid-search to adjust this parameters. 
    Note (last row): our approach is complementary with KD, assuming a pre-trained teacher is available beforehand. 
    \label{tab:comp_different_app}}
\end{table}

\mypar{Alternative teacher/student} 
In Table \ref{tab:comp_different_app}  we report the results obtained with the different distillation or co-training approaches depicted in Figure~\ref{fig:codist}. Other approaches are not effective off-the-shelf, except KD that requires a pre-trained teacher. 
Our approach is on par with KD (lower for ViT-L, better or ViT-H), and in the last row we show that it even provides a slight boost to combine KD with \ours.

\mypar{Impact of the \ours loss} 
The hyper-parameter $\lambda$ controls the tradeoff between the co-distillation loss and the cross-entropy classification loss. 
Setting $\lambda=1$ means that we have a regular supervised training setting, except that (i) we double the batch size by duplicating the image after data augmentation, and (ii)  stochastic depth selects different layers for each image copy. 

In Table~\ref{tab:submodel_performance}, we  measure the  impact of   $\lambda$, with all other hyper-parameters being fixed,
 for ViT-H trained at resolution $126\!\times\!126$ on Imagenet-1k.
We observe that the best ratio is to use an equal weighting of the \ours loss and the classic training loss. 
Using the \ours loss increases the performance  by 0.6\% Top-1 accuracy on Imagenet-val, which is the typical improvement that we observe for large models. %

\begin{table}
    \centering
    \scalebox{0.8}{
    \begin{tabular}{c@{\ \quad \ }c@{\ \quad \ }c@{\ \quad \ }c@{\ \quad \ }c@{\ \quad \ }c@{\ \quad \ }c}
    \toprule
        $\lambda$  & 0.1  &  0.3  & \cellcolor{blue!7}  \textbf{0.5}  & 0.7  & 0.9  & 1.0\\
        top1 accuracy      &  79.05 & 83.27 & \cellcolor{blue!7} \textbf{83.55} & 83.20 & 83.04  & 82.91 \\
        \bottomrule
    \end{tabular}}
        \vspace{-0.7em}
    \caption{Ablation of the parameter $\lambda$
controlling the weight of the co-distillation loss across submodels  (Imagenet1k-val, top1-acc).  Model ViT-H trained at resolution $126\!\times\!126$ on Imagenet-1k during 800 epochs. $\lambda=1.0$ corresponds a supervised baseline and $\lambda=0.5$ corresponds to \ours.
\label{tab:submodel_performance}}
\end{table}

\begin{table}
    \centering
    \scalebox{0.8}{
    \begin{tabular}{lcccccc}
        \toprule
        \multirow{2}{*}{Model} & \multirow{2}{*}{$\SD$} & \multirow{2}{*}{\ epochs\ } & \multicolumn{2}{c}{Baseline} & \multicolumn{2}{c}{+\ours}\\
         \cmidrule(lr){4-5} \cmidrule(lr){6-7}
         &  & & val & v2 & val & v2\\
         \midrule
         \multirow{2}{*}{CaiT-M12 \quad \ } & \multirow{2}{*}{0.20} & \pzo400 & \  83.2 \ & \ 72.9\  & \cellcolor{blue!7} \ 83.7 \ & \cellcolor{blue!7} \  \ 73.5 \ \ \\
                                   & & \pzo800 & 82.9 & 72.6 & \cellcolor{blue!7} 83.6 & \cellcolor{blue!7} 73.1 \\
        \\[-0.9em]

         \multirow{2}{*}{PiT-B} & \multirow{2}{*}{0.25} & \pzo400 & 83.8 & 73.6 & \cellcolor{blue!7} 84.1 &  \cellcolor{blue!7} 74.1\\
                                & & \pzo800 & 82.4 & 71.9 & \cellcolor{blue!7} 83.1 & \cellcolor{blue!7} 72.8  \\
        \\[-0.9em]
         \multirow{4}{*}{ViT-B} & \multirow{4}{*}{0.20} & \pzo400 & 83.1 & 72.6 & \cellcolor{blue!7} 83.2 & \cellcolor{blue!7} 73.1 \\
                                & & \pzo800 & 83.7 & 73.1 & \cellcolor{blue!7} 83.9 & \cellcolor{blue!7} 73.5 \\
                               & & 1200    & 83.3 & 72.8 & \cellcolor{blue!7} \_   & \cellcolor{blue!7} \_ \\
                               & & 1600    & 83.3 & 73.4 & \cellcolor{blue!7} \_   & \cellcolor{blue!7} \_ \\

        \\[-0.9em]
          \multirow{2}{*}{ViT-H}  & \multirow{2}{*}{0.60} & \pzo400 & 84.8 & 75.3 & \cellcolor{blue!7} 85.0 & \cellcolor{blue!7} 75.8 \\
                                 & & \pzo800 & 84.9 & 75.6 & \cellcolor{blue!7} 85.5 & \cellcolor{blue!7} 76.3 \\   
        \bottomrule
     \end{tabular}}
         \vspace{-0.7em}
    \caption{We compare ViT models trained with and without \ours on ImageNet-1k only with different number of epochs at resolution $224\!\times\!224$. 
    One can see that \ours is  more effective for larger models yielding higher values of the SD hyper-parameter $\SD$. 
    It avoids the early saturation or overfitting of the performance that we typically observe with the baseline when we increase the training time without re-adjusting hyper-parameters. See also Table~\ref{tab:submodel_performance} for a direct comparison with and without the co-distillation loss, and Table~\ref{tab:training_cost} for the corresponding training times per epoch. 
    }
    \label{tab:ablation_training_epochs}
\end{table}

\mypar{Number of training epochs}
In Table~\ref{tab:ablation_training_epochs} we compare results on Imagenet-1k-val and Imagenet-v2 different for architectures trained with and without \ours on Imagenet-1k only at resolution $224\!\times\!224$ with different number of epochs.
We observe less overfitting with \ours and longer training schedule. In particular, with bigger architecture like ViT-H, we observe  continuous improvement with a longer schedule where the baseline saturates.

\mypar{Training time}
In Table~\ref{tab:training_cost} we compare the training costs of \ours and DeiT-III. 
Thanks to our efficient stochastic depth formulation we maintain a similar memory peak during  training. 
For  bigger architectures the gap in training speed between \ours and the baseline is decreasing.

\begin{table}
    \centering
    \scalebox{0.8}{
    \begin{tabular}{lcccc}
    \toprule
         Training method & \ model \ & \ GPUs \ & Memory & Time  (min)   \\
                &       & used&  peak (GB) & by epoch \ \  \\
         \midrule
         \multirow{2}{*}{DeiT-III} & ViT-L & 32 & 21.4 & \pzo8 \\
                                   & ViT-H & 64 & 27.6 & 12 \\
                                                          \\ [-0.9em]
        \multirow{2}{*}{DeiT-III + ESD} & ViT-L & 32 & \cellcolor{red!7}15.1 &\cellcolor{red!7}\pzo9 \\
                                      & ViT-H & 64 & \cellcolor{red!7}15.2 &\cellcolor{red!7}11 \\
                                      \\ [-0.9em]
         \multirow{2}{*}{\ours (with ESD)\quad \ }  & ViT-L & 32 & \cellcolor{blue!7}26.9 & \cellcolor{blue!7}16 \\
                                  & ViT-H & 64 &\cellcolor{blue!7}25.0 &\cellcolor{blue!7}17 \\
        \bottomrule
    \end{tabular}}
    \vspace{-0.7em}
    \caption{Training times of different models trained at resolution $224\!\times\!224$ with batch size 2048 on Imagenet-1k with DeiT-III and our approach. 
    \ours uses our efficient stochastic depth (ESD), which amortizes the extra memory needed by \ours, especially for the largest models with  high stochastic depth values (0.45 for ViT-L, and 0.6 for ViT-H). 
    Timings are indicative and not representative of an optimized selection of the batch size. 
    \label{tab:training_cost}}
        \vspace{-0.5em}
\end{table}

\begin{table}
    \centering
    \scalebox{0.8}{
    \begin{tabular}{lccccc}
        \toprule
        \multirow{2}{*}{Model} &  \multirow{2}{*}{Resolution} & \multicolumn{2}{c}{Deit-III Baseline} & \multicolumn{2}{c}{+\ours}\\
                 \cmidrule(lr){3-4} \cmidrule(lr){5-6}
         &  & val & v2 & val & v2\\
         \midrule
         \multirow{3}{*}{ViT-B} & $128\times128$ & 83.5 & 73.4 & \cellcolor{blue!7} 83.8 & \cellcolor{blue!7} 74.0 \\
                                & $192\times192$ & 83.8 & 73.6 & \cellcolor{blue!7} 84.1 & \cellcolor{blue!7} 74.0 \\
                                & $224\times224$ & 83.7 & 73.1 & \cellcolor{blue!7} 83.9 & \cellcolor{blue!7} 73.5 \\
        \\[-0.9em]
          \multirow{3}{*}{ViT-L} & $128\times128$ & 84.5 & 74.7 & \cellcolor{blue!7} 85.1 & \cellcolor{blue!7} 75.5 \\
                                 & $192\times192$ & 84.9 & 75.1 & \cellcolor{blue!7} 85.2 & \cellcolor{blue!7} 75.7 \\
                                 & $224\times224$ & 84.5 & 75.0 & \cellcolor{blue!7} 84.9 & \cellcolor{blue!7} 75.6 \\
        \\[-0.9em]
          \multirow{3}{*}{ViT-H\quad \ \ \ \ } & \quad $126\times126$\quad \  & 85.1 & 75.6 & \cellcolor{blue!7}\ \ 85.6 \ \ & \cellcolor{blue!7} \ \ 76.4 \ \ \\
                                 & $182\times182$ & 85.1 & 75.9 & \cellcolor{blue!7} 85.7 & \cellcolor{blue!7} 76.6 \\
                                 & $224\times224$ & 84.9 & 75.6 & \cellcolor{blue!7} 85.5 & \cellcolor{blue!7} 76.3 \\   
        \bottomrule
     \end{tabular}}
    \vspace{-0.5em}
    \caption{Imagenet-1k val and v2 top-1 accuracy of ViT models trained with and without \ours for 800 epochs on Imagenet-1k at  different resolutions, followed by  finetuning for 20 epochs at resolution $224\!\times\!224$. 
    \label{tab:ablation_training_resolution}}
    \vspace{-0.5em}
\end{table}

\mypar{Resolution}
In Table~\ref{tab:ablation_training_resolution} we compare  different ViT architectures trained with and without \ours at different resolutions on Imagenet-1k. We fine-tune during 20 epochs at resolution $224 \times224$ before  evaluation at this resolution.
We observe that \ours gives  significant improvements across the different resolutions and models.

\begin{table}
    \centering
    \scalebox{0.8}{
    \begin{tabular}{l|cccc|cccc}
    \toprule
\quad	resol. $\rightarrow$  & 112 & 224 & 336	& 448 & 112 & 224 & 336	& 448 \\ 
	  \midrule
$\downarrow$ model    & \multicolumn{4}{c}{Imagenet-val} & \multicolumn{4}{c}{Imagenet-v2} \\
    \midrule
ViT-S & 78.0 & 83.1 & 84.6 & 85.2 & 66.6  & 73.7  & 75.1  & 76.3   \\
ViT-M & 80.6 & 85.0 & 86.0 & 86.3 & 69.6  & 76.0  & 76.8  & 77.2    \\
ViT-B & 82.8 & 86.3 & 86.9 & 87.4 & 72.1  & 77.0 & 77.9  & 78.3      \\
ViT-L & 85.4 & 87.5 & 88.1 & 88.3 & 75.7 & 79.1  & 79.8  & 80.0     \\
ViT-H & 86.2 & 88.0 &  - & - & 76.9  & 79.6  &  -  & -     \\
ViT-g & 86.5 & - &  - & -  & 77.3  &  - &  -  & -     \\
    \bottomrule
    \end{tabular}}
    
    \vspace{-0.5em}
    \caption{\textbf{Performance of  models at different resolutions}. We report the results obtained on Imagenet-val by models of different sizes pre-trained with \ours on Imagenet-21k and fine-tuned on Imagenet-1k. Training schedule: 270 epochs except ViT-g  (90 epochs). 
    Except for ViT-S, the results at resolution 336 and 448 were pre-trained on Imagenet-21k at resolution 224 for efficiency reasons. We have not fine-tuned ViT-H and ViT-g at large resolutions since these models are computationally expensive. 
\label{fig:resolution}}
\end{table}

\mypar{Imagenet-21k impact of layer-decay}
In Table~\ref{tab:ablation_21k} we compare different number of epochs for the pre-training on Imagenet-21k and the finetuning on Imagenet-1k with and without layer-decay. 
We observe that both layer-decay and long training bring  improvements with \ours.

\begin{table}
    \centering
    \scalebox{0.8}{
    \begin{tabular}{lcc|ccc}
         \toprule
         \multirow{2}{*}{Method} &  Long   & \ \ \ \ Layer \ \ \  & \multicolumn{3}{c}{Model}\\
                                 & \ \ Training\ \   & \ \ Decay\ \  & \ ViT-S \  & \  ViT-B \  & \ ViT-L \  \\
        \midrule
        \multirow{1}{*}{baseline\quad \ } 
                                 & \xmarkgr & \xmarkgr & 82.6 & 85.2 & 86.8\\
                                   \\[-0.9em]
        \multirow{4}{*}{\ours} & \xmarkgr & \xmarkgr &\cellcolor{blue!7}82.5 &\cellcolor{blue!7}85.8 & \cellcolor{blue!7}87.4\\
                               & \xmarkgr & \cmarkgr &\cellcolor{blue!7}82.7 & \cellcolor{blue!7}86.0 & \cellcolor{blue!7}87.5\\
                               & \cmarkgr & \xmarkgr &\cellcolor{blue!7}82.8 & \cellcolor{blue!7}86.0 & \cellcolor{blue!7}87.4\\
                               & \cmarkgr & \cmarkgr &\cellcolor{blue!7}83.1 & \cellcolor{blue!7}86.3 & \cellcolor{blue!7}87.5\\
        
        \bottomrule
    \end{tabular}}
    \vspace{-0.5em}
    \caption{We measure the impact of layer-decay during the finetuning on Imagenet-1k for models pre-trained on Imagenet-21k with \ours during 90 epochs (default) and 270 epochs (long training).  
    \label{tab:ablation_21k}}
\end{table}

\subsection{Comparisons with the state of the art}

\mypar{Imagenet-1k}
In Table~\ref{tab:mainres} we compare architectures trained with \ours with state-of-the-art results from the literature for this architecture. We observe that architectures trained with \ours are very competitive.
For instance, for ResNet-152, RegNetY-16GF, PiT-B we improve the best results reported in the literature by more than $0.9\%$ top-1 accuracy on Imagenet-1k-val.

\begin{table}

    \centering
    \scalebox{0.67}{
    \begin{tabular}{@{\ }l@{}c@{\ \ }c@{\ \ \ }r@{\ \ \ }r|cc@{\ }}
        \toprule
        Model        & nb params & throughput & FLOPs & Peak Mem & Top-1  & v2 \\
                      & ($\times 10^6$) & (im/s) & ($\times 10^9$) & (MB)\ \ \ \  & Acc.  & Acc. \\[3pt]

\toprule
\multicolumn{7}{c}{\textbf{``Traditional'' ConvNets}} \\[3pt]

     ResNet-50~\cite{He2016ResNet,wightman2021resnet} &  \pzo25.6    & 2587  &  4.1      & 2182 & 80.4  & 68.7 \\
    ResNet-101~\cite{He2016ResNet,wightman2021resnet}&  \pzo44.5    & 1586  &  7.9      & 2269 & 81.5  & 70.3  \\
    \rowcolor{blue!7}
    ResNet-101 -- \ours & \pzo44.5    & 1586  &  7.9      & 2269 & 82.1 & 70.8 \\
    ResNet-152~\cite{He2016ResNet,wightman2021resnet}&  \pzo60.2    &  1122 & 11.6      & 2359 & 82.0  & 70.6 \\
       \rowcolor{blue!7}
     ResNet-152 -- \ours &  \pzo60.2    &  1122 & 11.6      & 2359 &  83.1 & 72.1 \\
	 RegNetY-8GF~\cite{Radosavovic2020RegNet,wightman2021resnet}       & \pzo39.2  & 1158 & \tzo8.0 & 3939 & 82.2 & 71.1 \\
	 RegNetY-16GF~\cite{Radosavovic2020RegNet,Touvron2020TrainingDI}      & \pzo83.6  & \pzo714 & \dzo16.0 & 5204   & 82.9  & 72.4 \\

     \rowcolor{blue!7}
     RegNetY-16GF -- \ours~~\,   & \pzo83.6  & \pzo714 & \dzo16.0 & 5204   & 83.8  & 72.8 \\
     \\[-0.8em]

\multicolumn{7}{c}{\textbf{Vision Transformers derivatives}} \\ [5pt]
	Swin-T~\cite{liu2021swin} & \pzo28.3 & 1109\pzo & 4.5 & 3345 & 81.3 &  69.5 \\
    Swin-B~\cite{liu2021swin} & \pzo87.8  & 532 & 15.4 & 4695 & 83.5 &   \_ \\
    PiT-S~\cite{Heo2021RethinkingSD} & \pzo23.5 & 1809\pzo & 2.9 &  3293 & 80.9 & \_\\
    \rowcolor{blue!7}
     PiT-S -- \ours & \pzo23.5 & 1809\pzo & 2.9 &  3293 & 81.3 & 69.7 \\
     
    PiT-B~\cite{Heo2021RethinkingSD} & \pzo73.8 & 615 & 12.5  &  7564 & 82.0 & \_\\
    \rowcolor{blue!7}
     PiT-B -- \ours & \pzo73.8 & 615 & 12.5  &  7564 & 84.1 &74.1 \\
     \\[-0.8em]
    
 \multicolumn{7}{c}{\textbf{Vanilla Vision Transformers}} \\[3pt]

    ViT-S ~\cite{Touvron2020TrainingDI} & \pzo22.0  & 1891\pzo & \tzo4.6 & 987 & 79.8 & 68.1  \\
    ViT-S -- Deit III~\cite{touvron2022deitIII} & \pzo22.0  & 1891\pzo & \tzo4.6 & 987 & 81.4 &  70.5 \\
        \rowcolor{blue!7}
    ViT-S -- \ours & \pzo22.0  & 1891\pzo & \tzo4.6 & 987 & 81.5 &  70.8 \\

    ViT-B~\cite{dosovitskiy2020image}  & \pzo86.6  & 831  & \dzo17.5 & 2078 & 77.9 &  --\\
    ViT-B~-- DeiT\cite{Touvron2020TrainingDI}  & \pzo86.6  & 831  & \dzo17.5 & 2078 & 81.8 & 71.5 \\
    ViT-B~-- DeiT/distilled & \pzo86.6  & 831  & \dzo17.5 & 2078 & 83.4 & 73.2 \\
    ViT-B -- Deit III~\cite{touvron2022deitIII}  & \pzo86.6  & 831  & \dzo17.5 & 2078 & 83.8 &  73.6\\
    
    \rowcolor{blue!7}
    ViT-B -- \ours & \pzo86.6  & 831  & \dzo17.5 & 2078 & 84.2 & 74.2 \\
    
    ViT-L -- Deit III~\cite{touvron2022deitIII}  & 304.4 & 277 & 61.6 & 3789 & 84.9 & 75.1    \\
    \rowcolor{blue!7}
    ViT-L -- \ours & 304.4 & 277 & 61.6 & 3789 & 85.3 &  75.5   \\
    ViT-H -- Deit III~\cite{touvron2022deitIII}  & 632.1 & 112 & 167.4 & 6984 & 85.2 & 75.9     \\

    \rowcolor{blue!7}
    ViT-H -- \ours  & 632.1 & 112 & 167.4 & 6984 & 85.7 &  76.6 \\

    \bottomrule
    \end{tabular}}
    \vspace{-0.5em}
  \caption{
\textbf{Classification with Imagenet1k training.} 
We compare with models trained on Imagenet-1k only at resolution $224\!\times\!224$ without self-supervised pre-training (see supp. material for a comparison).
We report Top-1 accuracy on Imagenet-1k-val and Imagenet-v2 with different measures of complexity: throughput, FLOPs, number of parameters and peak memory usage. 
The throughput and peak memory are measured on a single V100-32GB GPU with batch size fixed to 256 and mixed precision. 
\label{tab:mainres}}
\end{table}

\mypar{Imagenet-21k} In Table~\ref{tab:mainres_22k} we compare ViT models pre-trained with \ours on Imagenet-21k and finetuned with \ours on Imagenet-1k with other architectures and our baseline DeiT-III.
Our results with vanilla ViT outperform the baseline and are competitive with  recent architectures. 

\begin{table}
    \centering
\scalebox{0.66}{
    \begin{tabular}{@{\ }l@{}c@{\ \ }c@{\ \ \ }r@{\ \ \ }r|cc@{\ }}
        \toprule
        Architecture        & nb params & throughput & FLOPs & peak mem & \multicolumn{2}{c}{top1 acc.}  \\
                      & ($\times 10^6$) & (im/s) & ($\times 10^9$) & (MB)\ \ \ \  & val  & v2 \\[3pt]

\toprule
\multicolumn{7}{c}{\textbf{Convnets}}  \\[3pt]

EfficientNetV2-M~\cite{Tan2021EfficientNetV2SM} & \pzo54.1 & 312 & 25.0 & 7127 & 86.2 & 75.9 \\
EfficientNetV2-L~\cite{Tan2021EfficientNetV2SM} & 118.5 & 179 & 53.0 & 9540 & 86.8 & 76.9 \\
    \\[-0.7em]  

    ResNet-152~\cite{He2016ResNet,wightman2021resnet}&  \pzo60.2    &  1122\pzo & 11.6      & 2359 & 82.0  & 70.6 \\
\rowcolor{blue!7}
    ResNet-152 -- \ours & \pzo 60.2    &  1122\pzo & 11.6      & 2359 & 83.1 & 73.1 \\
         \\[-0.9em]  

\rowcolor{blue!7}
	 RegnetY16GF -- \ours \quad~   & \pzo83.6  & 714 & \dzo16.0 & 5204   & 84.2  & 74.7 \\

    \\[-0.9em]

\rowcolor{blue!7}
    ConvNeXt-S-- \ours & \pzo 50.2 & 783 & \pzo8.7 &2218 &  85.2 &  76.0 \\     
    ConvNeXt-B \cite{Liu2022convnext}& \pzo88.6 & 563 & 15.4 &3029 &  85.8 & 75.6 \\     
\rowcolor{blue!7}
    ConvNeXt-B -- \ours & \pzo88.6 & 563 & 15.4 &3029 &  85.8 &  76.9 \\     
    
    ConvNeXt-L \cite{Liu2022convnext} & 197.8 & 344 & 34.4 & 4865 &  86.6 & 76.6\\     
    
    ConvNeXt-XL  \cite{Liu2022convnext}& 350.2 & 241 & 60.9 & 6951 &  87.0 & 77.0\\     
    \\[-0.9em]

\multicolumn{7}{c}{\textbf{Transformers variations}} \\ [3pt]
     Swin-B~\cite{liu2021swin} & \pzo87.8 & 532 & 15.4 & 4695 &  85.2 & 74.6 \\
\rowcolor{blue!7}
     Swin-B -- \ours  & \pzo87.8 & 532 & 15.4 & 4695 & 86.2 &  77.2  \\
     
     Swin-L~\cite{liu2021swin} & 196.5 & 337 & 34.5 & 7350 &  86.3 & 76.3 \\
\rowcolor{blue!7}
     Swin-L -- \ours  & 196.5 & 337 & 34.5 & 7350 &  87.1 &  78.1  \\
         \\[-0.9em]  

\rowcolor{blue!7}
         PiT-B -- \ours~\cite{Heo2021RethinkingSD}       & \pzo73.8     & 615 & 12.5   & 7564 &  \cellcolor{blue!7}85.8         &  76.8 \\
            \\[-0.9em]  
\rowcolor{blue!7}
         XCiT-S12 -- \ours~\cite{el2021xcit}              &  \pzo26.2    &  1373 &  4.9 & 1330   &  \cellcolor{blue!7}84.2                   &  74.9 \\
\rowcolor{blue!7}
         XCiT-M24 -- \ours~\cite{el2021xcit}              &  \pzo84.4    & 553  &  16.2  & 2010    &  \cellcolor{blue!7}86.5                &  78.0 \\
\rowcolor{blue!7}
         XCiT-L24 -- \ours~\cite{el2021xcit}              &  189.0       & 334  &  36.1  & 3315   &  \cellcolor{blue!7}87.2                 &  77.8 \\
    \\[-0.8em]  

\multicolumn{7}{c}{\textbf{Vanilla Vision Transformers}~\cite{dosovitskiy2020image,touvron2022deitIII} } \\ [3pt]

\rowcolor{blue!7}
ViT-S -- \ours & \pzo22.0  & 1891\pzo  & \tzo4.6 & \pzo987 & 83.1 & 73.7 \\
\rowcolor{blue!7}
ViT-M -- \ours & \pzo39.0  & 1307\pzo  & \tzo8.0 & 1322 & 85.0 & 76.0 \\
ViT-B -- DeiT-III  & \pzo86.6  & 831  & \dzo17.6 & 2078 & 85.7 & 76.5 \\
\rowcolor{blue!7}
ViT-B -- \ours & \pzo86.6  & 831  & \dzo17.6 & 2078 & 86.3 & 77.0 \\

ViT-L -- DeiT-III & 304.4 & 277 & 61.6 & 3789  & 87.0 &  78.6 \\
\rowcolor{blue!7}
ViT-L -- \ours & 304.4 & 277 & 61.6 & 3789  & 87.5 &  79.1 \\

ViT-H -- DeiT-III & 632.1 & 112 & 167.4 & 6984 & 87.2 & 79.2 \\
 
\rowcolor{blue!7}
ViT-H -- \ours & 632.1 & 112 & 167.4 & 6984 & 88.0 & 80.0 \\
    \bottomrule
    \end{tabular}}
    \vspace{-0.7em}
\caption{\textbf{Classification with ImageNet-21k pretraining.} 
We report top-1 accuracy on the validation set of Imagenet1k and Imagenet-V2 with different measures of complexity. The peak memory usage is measured on a single V100-32GB GPU with batch size fixed to 256 and mixed precision. For Swin-L the memory peak is an estimation since we decreased the batch size to 128 to fit in memory.
All models are evaluated at resolution 224 except EfficientNetV2 that use resolution 480. 
ViT are pretrained wit a $\times$3 schedule, comparable to that used in the best DeiT-III baseline (270 vs.\ 240 epochs). All other \ours models are pretrained during 90 epochs on Imagenet21k, with 50 epochs of fine-tuning. The $\tau$ hyper-parameter is set per model based on prior choices or  best guess based on model size. \label{tab:mainres_22k}}
    \vspace{-0.4em}
\end{table}

\begin{figure}
    \begin{minipage}{0.48 \linewidth}
    ~ \hfill \quad \footnotesize Imagenet-1k \hfill ~    

    \includegraphics[height=0.95\linewidth]{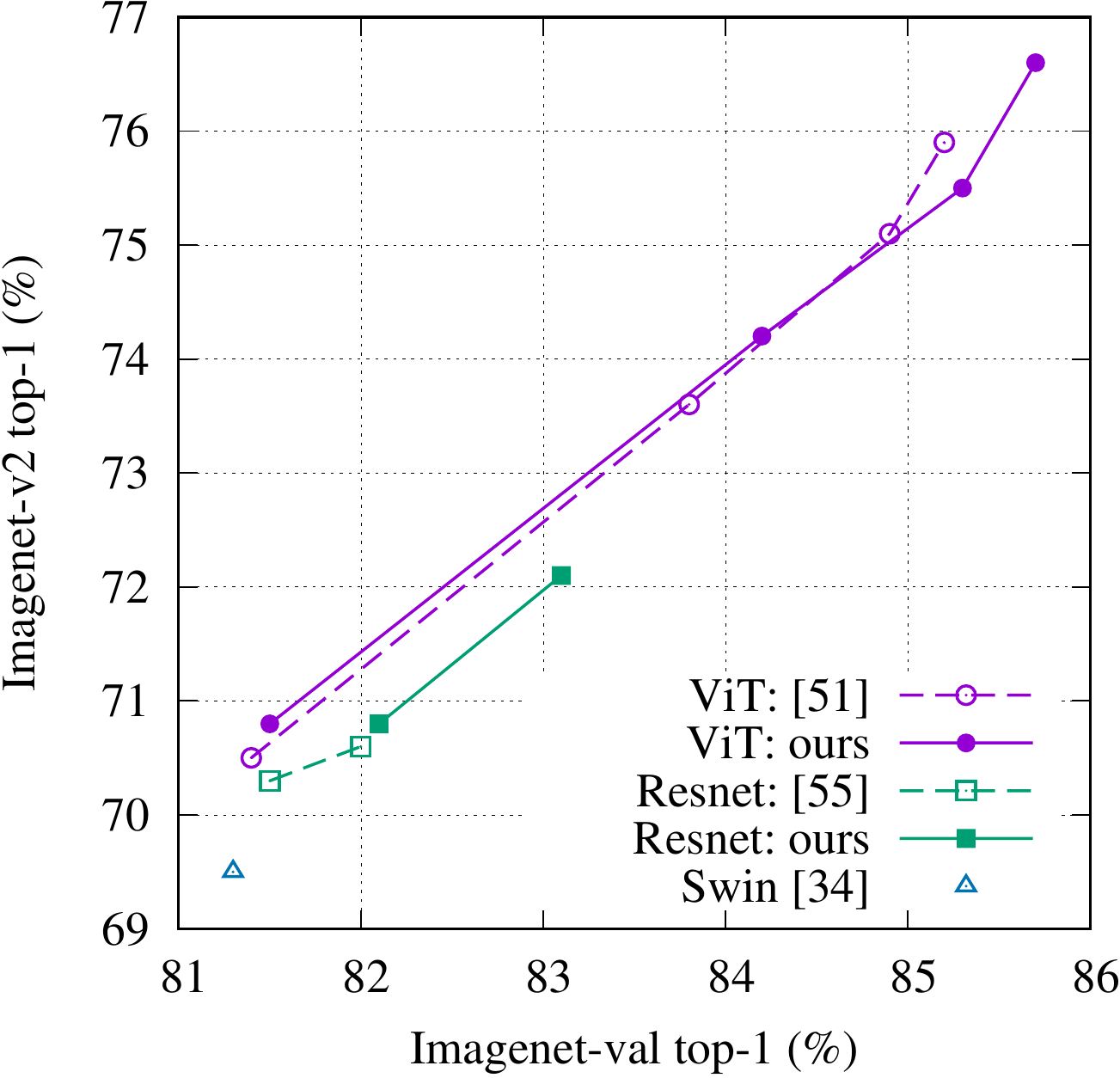}
    \end{minipage}
    \hfill
    \begin{minipage}{0.48 \linewidth}
    ~ \hfill \footnotesize Imagenet-21k \hfill ~
    
    \includegraphics[trim={30pt 0 0 0},clip,height=0.95\linewidth]{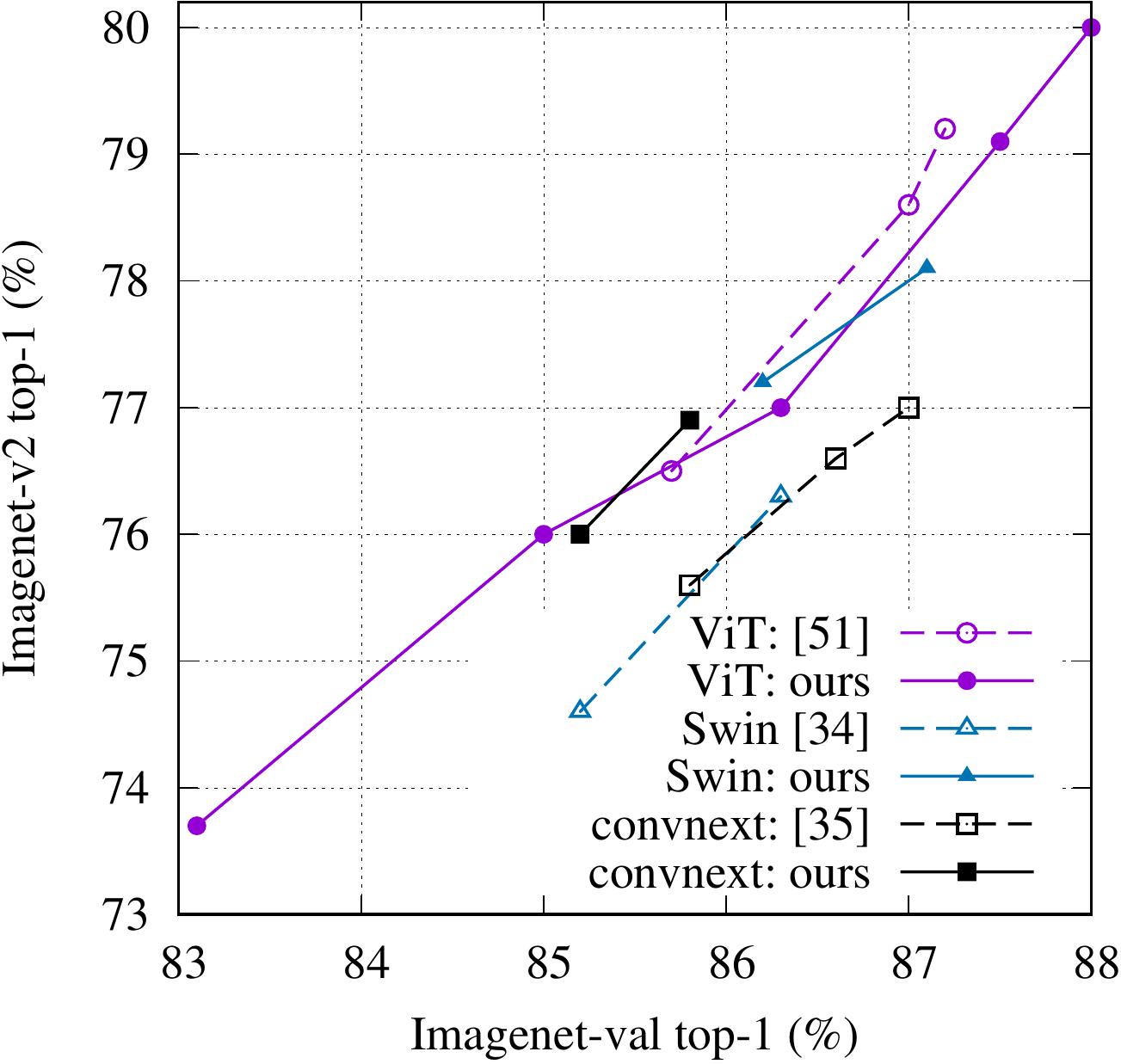}
    \end{minipage}
    \vspace{-0.5em}
    \caption{Overfitting measurement: top-1 accuracy on Imagenet-val vs.\ Imagenet-v2 for models in Tables~\ref{tab:mainres} and~\ref{tab:mainres_22k} pre-trained on Imagenet-1k and Imagenet-21k, respectively. Our cosub ViTs (plain lines and points) do not overfit more than DeiT-III~\cite{touvron2022deitIII} overall. Our Imagenet-21k results for Swin and Convnext generalize much better on v2 than the original models. 
    %
    \label{fig:imagenet_v2}
    }
    \vspace{-2em}
\end{figure}

\mypar{Overfitting analysis}
As recommended in prior works~\cite{wightman2021resnet,touvron2022deitIII} we perform an overfitting analysis. We evaluate our models trained with codist on Imagenet-1k val and Imagenet-v2~\cite{Recht2019Imagenetv2}. The results are reported in Figure~\ref{fig:imagenet_v2}. For ViT, we observe that \ours does not overfit more than the DeiT-III baseline~\cite{touvron2022deitIII}. Our results with other architectures in Table~\ref{tab:mainres_22k} concur with that observation: our results are comparatively stronger on Imagenet-v2 than those reported in the literature for the exact same models. 

\subsection{Downstream tasks}

\mypar{Semantic segmentation}
First we evaluate our ViT models pre-trained on Imagenet with \ours for semantic segmentation on the  ADE20k dataset~\cite{Zhou2017ScenePT}. 
ADE20k consists of 20k training and 5k validation images with labels over 150 categories. 
We adopt the training schedule from Swin:~160k iterations with UperNet~\cite{xiao2018unified}. 
At test time we evaluate with a single scale and multi-scale.
Our UperNet implementation is the same as in  DeiT III~\cite{touvron2022deitIII}.
Our results are reported in Table~\ref{tab:sem_seg}.
We observe that vanilla ViTs trained with \ours outperform our baseline DeiT-III but also have better FLOPs-accuracy trade-offs than recent architectures. %

\begin{table}
        \centering
        \scalebox{0.73}{
        \begin{tabular}{lcccc}
        \toprule
             \multirow{2}{*}{Backbone}  
              & \#params \ \ & \ \  FLOPs \ \ & Single-scale & Multi-scale  \\
             &  ($\times 10^6$) & ($\times 10^9$) & mIoU  & mIoU \\
            \midrule 
             DeiT-S  &  \pzo52.0 & 1099 & \_ & 44.0 \\

             Swin-T    & \pzo59.9 & \pzo945 & 44.5 & 46.1 \\

            ViT-S -- DeiT-III  & \pzo41.7 & \pzo588 & 45.6 & 46.8\\
            \rowcolor{blue!7}
            ViT-S -- \ours & \pzo41.7 & \pzo588 & \textbf{47.0} & \textbf{48.0}\\
            \multicolumn{5}{c}{}\\[-0.7em]

            PatchConvNet-B60 & 140.6 & 1258  & 48.1  & 48.6 \\

             PatchConvNet-B120  & 229.8 & 1550  & 49.4  & 50.3\\
             MAE ViT-B  & 127.7 & 1283 & 48.1 &  \_ \\
             Swin-B   & 121.0 & 1188  & 48.1 & 49.7 \\
            ViT-B -- DeiT-III  & 127.7 & 1283 & 49.3 &  50.2\\
            \rowcolor{blue!7}
            ViT-B -- \ours & 127.7 & 1283 & 49.3 &  49.9\\
            ViT-L -- DeiT-III  & 353.6 & 2231 & 51.5 & 52.0  \\    
            \rowcolor{blue!7}
            ViT-L -- \ours & 353.6 & 2231 & \textbf{52.5} & \textbf{53.1}  \\  
            \multicolumn{5}{c}{}\\[-0.7em]

            PatchConvNet--B60$^\dagger$  & 140.6 & 1258  & 50.5  & 51.1 \\

            Swin-B$^\dagger$ ($640\times 640$)  & 121.0 & 1841  & 50.0 & 51.6 \\
            ViT-B -- DeiT-III $^\dagger$  & 127.7 & 1283 & 53.4 & 54.1\\
            \rowcolor{blue!7}
            ViT-B -- cosub$^\dagger$  & 127.7 & 1283 & 53.7  & 54.7 \\
            PatchConvNet-L120$^\dagger$  & 383.7 & 2086  &  52.2 & 52.9   \\
            Swin-L$^\dagger$ ($640\times 640$)  & 234.0 & 3230  & \_ & 53.5 \\
            ViT-L -- DeiT-III $^\dagger$  & 353.6 & 2231 &  54.6 & 55.6 \\
            \rowcolor{blue!7}
            ViT-L -- cosub$^\dagger$  & 353.6 & 2231 & \textbf{55.7} & \textbf{56.3}  \\
        \bottomrule     
        \end{tabular}
        } 
            \vspace{-0.5em}
          \caption{\textbf{ADE20k semantic segmentation} performance using UperNet \cite{xiao2018unified},  in comparable setting as prior works~\cite{Dong2021CSWinTA,el2021xcit,liu2021swin}. 
          All models are pre-trained on Imagenet-1k,  except bottom models identified with $^\dagger$, which are pre-trained on Imagenet-21k. By default the finetuning resolution on ADE20k is $512\!\times\!512$ except when mentioned otherwise (for Swin).
       \label{tab:sem_seg}}
\end{table}

\mypar{Transfer learning}
We now measure how the performance improvements observed with \ours  translate to other classification problems. For this purpose, we performed transfer learning on the six different datasets used in DeiT-III. 
Our results are reported Table~\ref{tab:sota_tl}. 
Our pre-trained and finetuned models with \ours generally improve the baseline. The gains are overall more significant on the more challenging datasets like iNaturalist 2018 and iNaturalist 2019.

\begin{table}

    \centering
    \scalebox{0.74}{
    \begin{tabular}{l@{}c@{\ \ }c@{\ \ }cccc}
    \toprule
    Model    & Cifar-10 & Cifar-100  & Flowers & Cars & iNat-18 & iNat-19  \\
    \midrule                                                                          

    ViT-S -- DeiT-III  & 98.9 & 90.6 & 96.4 & 89.9 & 67.1 & 72.7 \\
    ViT-B -- DeiT-III  & 99.3 & 92.5 & 98.6 & 93.4 & 73.6 & 78.0\\ 
    ViT-L -- DeiT-III  & 99.3 & 93.4 & \textbf{98.9} & \textbf{94.5} & 75.6 & 79.3 \\ 
    \rowcolor{blue!7}
    ViT-S -- \ours & 99.1 & 91.7 & 97.4 & 93.0 & 70.1 & 75.6 \\
    \rowcolor{blue!7}    
    ViT-B -- \ours & 99.1 & 92.6 & 98.4 & 93.5 & 74.1 & 78.1\\ 
    \rowcolor{blue!7}
    ViT-L -- \ours & \textbf{99.4} & \textbf{93.5} & 98.8 & \textbf{94.5} & \textbf{76.2} & \textbf{80.1}  \\
    \bottomrule
    \end{tabular}}
    \vspace{-0.7em}
        \caption{
        ViT models pre-trained with \ours or DeiT-III on Imagenet-1k and finetuned on six different target datasets. 
        We note that for small datasets (CIFAR, Flowers, and Cars) our approach is useful for small models but neutral for larger models.
        The gains are more significant when transferring to the larger iNaturalist-18 and iNaturalist-19 datasets. 
    \label{tab:sota_tl}}
    \vspace{-0.57em}
\end{table} %

\section{Conclusion}

Co-training submodels (\ours) is an effective way to improve existing deep residual networks.
It is straightforward to implement, just involving a few lines of code. %
It does not need a pre-trained teacher, and it only maintains a single set of weights for the model. 
Extensive experimental results on image classification, transfer learning and semantic segmentation show that \ours is overall extremely effective.  
It works off-the-shelf and improves the state of the art for various network architectures, including convnets like Regnet. %

\clearpage

{\small
\bibliographystyle{ieee_fullname}
\bibliography{egbib}
}

\clearpage

\appendix
\clearpage
\counterwithin{figure}{section}
\counterwithin{table}{section}
\counterwithin{equation}{section}

\pagenumbering{Roman}  

\twocolumn 
[
\begin{center}
{\Large \bf \inserttitle \\ \vspace{0.5cm} \large supplemental material \par \vspace{2em}
}
\end{center}
]

\section{Training details}
\label{app:training_details}

\subsection{Hyper-parameters}

We use by default the DeiT-III training procedure from Touvron \etal~\cite{touvron2022deitIII}, which uses separate recipes for Imagenet21k and Imagenet1k. 
The most noticeable difference is the loss, which is by default a binary cross-entropy when training on Imagenet1k, versus a cross-entropy when pre-training with Imagenet21k and fine-tuning with Imagenet1k.  
We depart from the choices of DeiT-III as follows. First, we systematically set the weight decay to 0.02, independent of whether we pre-train on Imagenet21k or train or fine-tune on Imagenet1k. This does not change significantly the results with \ours. Our long schedule on Imagenet21k is systematically set to 270 epochs. 

\paragraph{Batch size and learning rate. } 
Second, the default batch size is by default set to 2048. However we need to reduce it to limit the memory consumption for larger models or higher resolution.  In particular, during the fine-tuning stage we adjust the learning accordingly and employ a square-root scaling rule compatible with AdamW~\cite{Loshchilov2017AdamW}: we fix the base learning rate as 
\begin{align}
    \mathrm{LR}_{\mathrm{train}}= 10^{-3} \sqrt{\frac{\mathrm{BS}}{2048}}, 
\end{align}
for pre-training on Imagenet21k with 90 epochs or training on Imagenet1k from scratch (400 or 800 epochs). When fine-tuning from Imagenet21k to Imagenet1k, we divide by 10 the base learning rate when starting from an existing model. Therefore we set 
\begin{align}
\mathrm{LR}_{\mathrm{finetune}}= 10^{-4} \sqrt{\frac{\mathrm{BS}}{2048}}  \times{C_\mathrm{LD}}, 
\end{align}
where we set the constant $C_\mathrm{LD}$\,=1  by default. This constant is modified when using LayerDecay~\cite{clark2020electra}, see below.

\paragraph{LayerDecay} is used in fine-tuning stages of recent self-supervised methods~\cite{bao2021beit,He2021MaskedAA}. It decreases the learning rate in a geometrically decreasing manner: the learning for each block $l$ is given by $\mathrm{LR}(l)=\mathrm{LD}^{l-L}$, where $\mathrm{LR}$ is the layer-wise decay factor, and $L$ is the total number of blocks of the network (e.g., 32 for a ViT-H). Hence the LR of the all layers is affected, except those in the final block. 

\begin{table}[t]
    \centering
    \scalebox{0.8}{
    \begin{tabular}{l@{\quad\quad}cccccc}
    \toprule
model  & ViT-S & ViT-M & ViT-B & ViT-L & ViT-H  \\
\midrule
$\tau$: Imnet1k  train              & 0.05 &  0.1 & 0.2  & 0.45 & 0.6 \\
$\tau$: Imnet21k pre-train          & 0.05 & 0.05 & 0.1  & 0.3 & 0.5  \\
LayerDecay                          & 0.7 & 0.75 & 0.75 & 0.8 & 0.85\\
    \bottomrule
    \end{tabular}}
    \vspace{-0.5em}
    \caption{Hyper-parameters that are set depending on the model size. 
    \label{tab:hparams}}
\end{table}

BeiT sets the LayerDecay parameter to LD\,=\,0.65 or 0.75 based on the model size, like OmniMAE~\cite{girdhar2022omnimae}. MAE~\cite{He2021MaskedAA} sets LD to 0.75. Another recent work uses 0.65~\cite{assran2022masked}. 
From our own preliminary experiments, we concur with the choice of Bao \etal~\cite{bao2021beit}, who adjust this parameter depending on the model size when fine-tuning from Imagenet21k to Imagenet1k. 
Hence we gradually increase the LD value from smaller to larger models. Table~\ref{tab:hparams} gives the value of this parameter for each model size. %

We set $C_\mathrm{LD}=2$ if we use LayerDecay in the fine-tuning stage: if a given learning rate was initially optimized without LayerDecay, it is necessary to compensate the overall reduction of updates. This was suggested by Bao \etal~\cite{bao2021beit}, but without any guideline on how to adjust the learning rate.   
From a few experiments, we notice that the simple formulaic choice of multiplying the learning by a constant 2 generally gives reasonable results. They could likely be further improved by further cross-validation, however this would require a much heavier set of experiments per model size.

\paragraph{Hyper-parameter $\tau$.} The other hyper-parameter that depends  on the model size is the so-call drop-path rate $\tau$ associated with stochastic depth~\cite{Huang2016DeepNW}. We report these values in Table~\ref{tab:hparams}. 
In particular, the value $\tau$ is inherently intertwined with our approach, as it is used to instantiate submodels.
A value of $\tau$ means that we instantiate two identical submodels, which zeroes the cross-entropy and therefore cancels our method \ours. More generally, higher values of $\tau$ provide submodels that have less layers in common. Therefore, we observe that it is beneficial to increase $\tau$ compared to the values suggested in the DeiT-III training method, especially for the smaller model Vit-S for which $\tau$ was initially set to 0. 
We further increase this rate by +0.05 when more regularization is needed, i.e., when pre-training during 270 epochs on Imagenet21k or for large resolutions.

\subsection{Transfer Learning datasets}

For the transfer learning tasks we fine-tune our ViT models pre-trained at resolution $224 \times 224$ on ImageNet-1k only with \ours on the 6 transfer learning datasets used in Touvron et al.~\cite{touvron2022deitIII}.
In Table~\ref{tab:dataset}, we give the characteristics of these datasets and corresponding references.

\begin{table}
\centering
\scalebox{0.8}{
\begin{tabular}{l@{\quad\quad\quad}r@{\quad\quad\quad}r@{\quad\quad\quad}r}
\toprule
Dataset & Train size & Test size & \#classes   \\
\midrule
iNaturalist 2018~\cite{Horn2018INaturalist}& 437,513   & 24,426 & 8,142 \\ 
iNaturalist 2019~\cite{Horn2019INaturalist}& 265,240   & 3,003  & 1,010  \\ 
Flowers-102~\cite{Nilsback08}& 2,040 & 6,149 & 102  \\ 
Stanford Cars~\cite{Cars2013}& 8,144 & 8,041 & 196  \\  
CIFAR-100~\cite{Krizhevsky2009LearningML}  & 50,000    & 10,000 & 100   \\ 
CIFAR-10~\cite{Krizhevsky2009LearningML}  & 50,000    & 10,000 & 10   \\ 
\bottomrule
\end{tabular}}%

\vspace{-0.5em}
\caption{Datasets used for our different transfer-learning tasks.  \label{tab:dataset}}
\end{table}

\section{Supplemental experiments}
\label{app:exp}

\subsection{New baselines for models of the literature} 

Table~\ref{tab:comp_archi_imnet21k} provides the results we obtained with minimal adjustments to our training recipe based on DeiT-III combined with submodel co-training. The only parameter that absolutely needs to be re-adjusted is $\tau$. For this purpose, we first tried existing parameter setting from the literature when existing ($\tau=0$ disables \ours therefore we use 0.05 in such cases). Otherwise we make a guess based on the model size and adjusts by step of 0.1 when the training curve exihbits some overfitting. This minimum hyper-parameter modification with a coarse step for $\tau$ is most likely suboptimal and could certainly be improved, but it would require much more compute capacity to optimize it with a proper cross-validation for each model. 

As one can see, our method improves the results for most of the models from the literature that we tested, therefore we hope that they could serve as improved baselines when comparing architectures. We also notice that our results on Imagenet-v2 are generally better than those reported in the literature. For instance our ConvNext-B training is comparable to that the original paper on Imagenet-val, but \ours's result on Imagenet-v2 is more than 1\% higher, which suggests that our training recipe overfits significantly less. 

As a disclaimer, we believe that Table~\ref{tab:comp_archi_imnet21k} should not be used as a way to compare the merits of architectures, since our training procedure may favor certain of them. As importantly and as mentioned above, we have put a minimal effort to obtained these results and it is highly likely that our hyper-parameters are very suboptimal for some models. 

\begin{table}
    \centering
    \scalebox{0.8}{
    \begin{tabular}{@{\ }l@{}cclccc@{\ }}
    \toprule
               & params  & FLOPS & previous   & \multicolumn{2}{c}{\ours acc.}\\
         Model & (M) & ($\times10^9)$ & top1 acc. &  -val &  -v2 \\
         \midrule
         ResNet-152~\cite{He2016ResNet}          & \pzo60 & 11.6    & 82.0$^\mathrm{(1k)}$\cite{wightman2021resnet}  &  \cellcolor{blue!7}83.1 &  73.1 \\
         RegNet-16GF\cite{Radosavovic2020RegNet}& \pzo84 & 16.0    & 82.2$^\mathrm{(1k)}$\cite{wightman2021resnet}  &  \cellcolor{blue!7}84.2  &  74.7 \\
         PiT-B~-distilled\cite{Heo2021RethinkingSD}    & \pzo74 & 12.5    & 84.5$^\mathrm{(1k)}$  &  \cellcolor{blue!7}85.8                    &  76.8 \\
         ConvNext-S~\cite{Liu2022convnext}       & \pzo50     & \pzo8.7 & 83.1$^\mathrm{(1k)}$ &  \cellcolor{blue!7}85.2                       &  76.0 \\
         ConvNext-B~\cite{Liu2022convnext}       & \pzo89     & 15.4 &   85.8$^\mathrm{(21k)}$   &  \cellcolor{blue!7}85.8                     &  76.9 \\
         XCiT-S12~\cite{el2021xcit}              &  \pzo26    &  \pzo4.9 &  83.3$^\mathrm{(1k)}$  &  \cellcolor{blue!7}84.2                          &  74.9 \\ 
         XCiT-M24~\cite{el2021xcit}              &  \pzo84    &  16.2   &  84.3$^\mathrm{(1k)}$  &  \cellcolor{blue!7}86.5                     &  77.9 \\
         XCiT-L24~\cite{el2021xcit}              &  189   &  36.1     &  84.9$^\mathrm{(1k)}$  &  \cellcolor{blue!7}87.2                       &  77.8 \\
         Swin-B~\cite{liu2021swin}               &   \pzo88    & 15.4  &  85.2$^\mathrm{(21k)}$  &  \cellcolor{blue!7}86.2                     &  77.2 \\
         Swin-L~\cite{liu2021swin}               &   197   & 34.5  &  86.3$^\mathrm{(21k)}$ &  \cellcolor{blue!7}87.1                          &  78.1 \\
\bottomrule

    \end{tabular}}
    \vspace{-0.5em}
    \caption{\textbf{New baselines for multiple architectures at resolution 224:} trained with \ours on Imagenet21k data. We adopt the same pre-training recipe (90 epochs of Imnet21k pretraining and 50 epochs of fine-tuning) and adjust the $\tau$ parameter per architecture based on prior choices or best guess based on model size. 
    These choices could most likely be improved by cross-validated grid search. We report good results reported in literature (in some cases obtained by training on Imagenet-train only:$^\mathrm{(1k)}$)   
    \label{tab:comp_archi_imnet21k}}
\end{table}

\subsection{Trade-offs between resolution and model size}

\begin{figure*}
    \begin{minipage}[b]{1.0\linewidth}
    \includegraphics[height=0.25\linewidth]{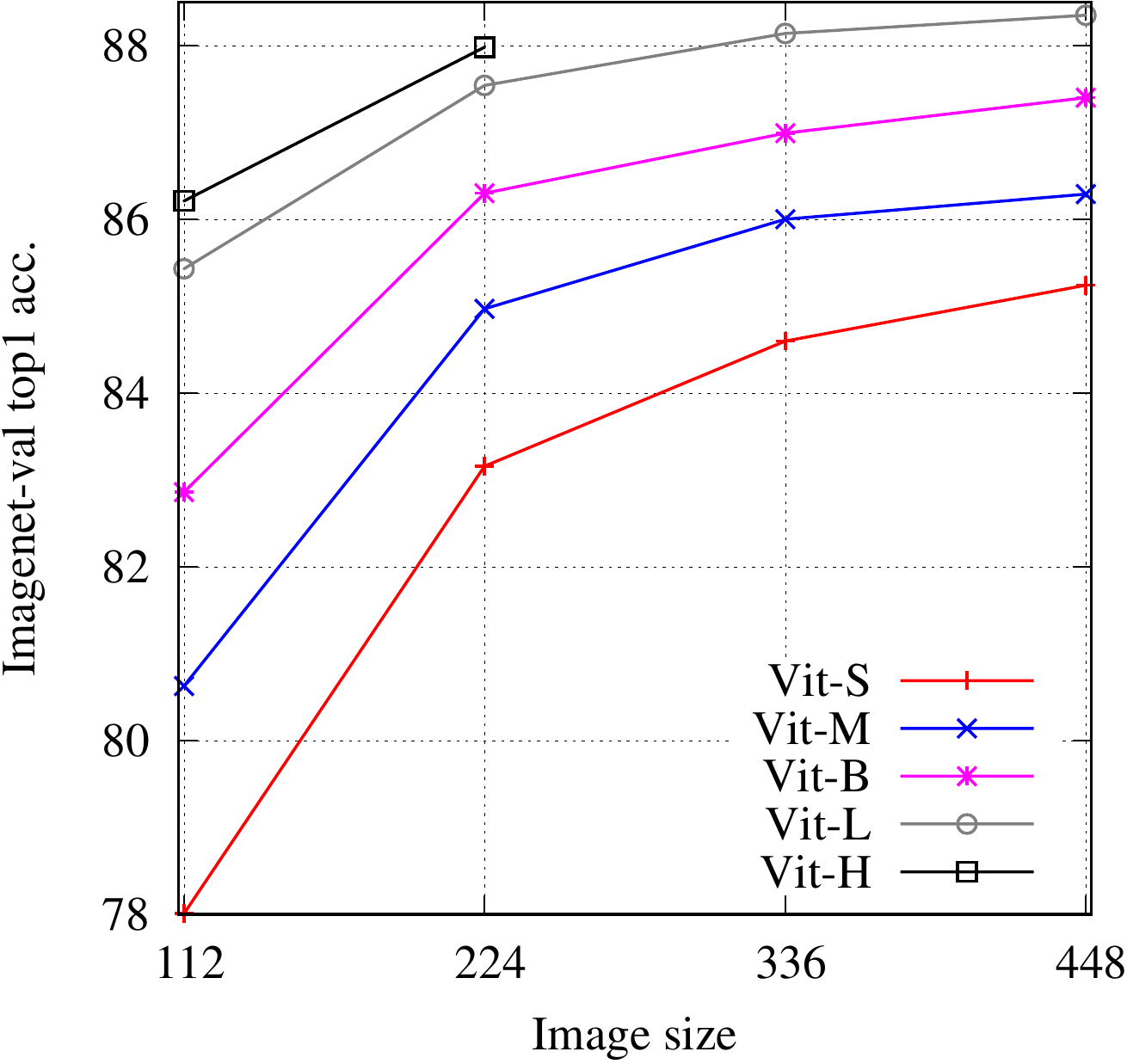}
    \hfill
    \includegraphics[height=0.25\linewidth,trim={5em 0 0 0},clip]{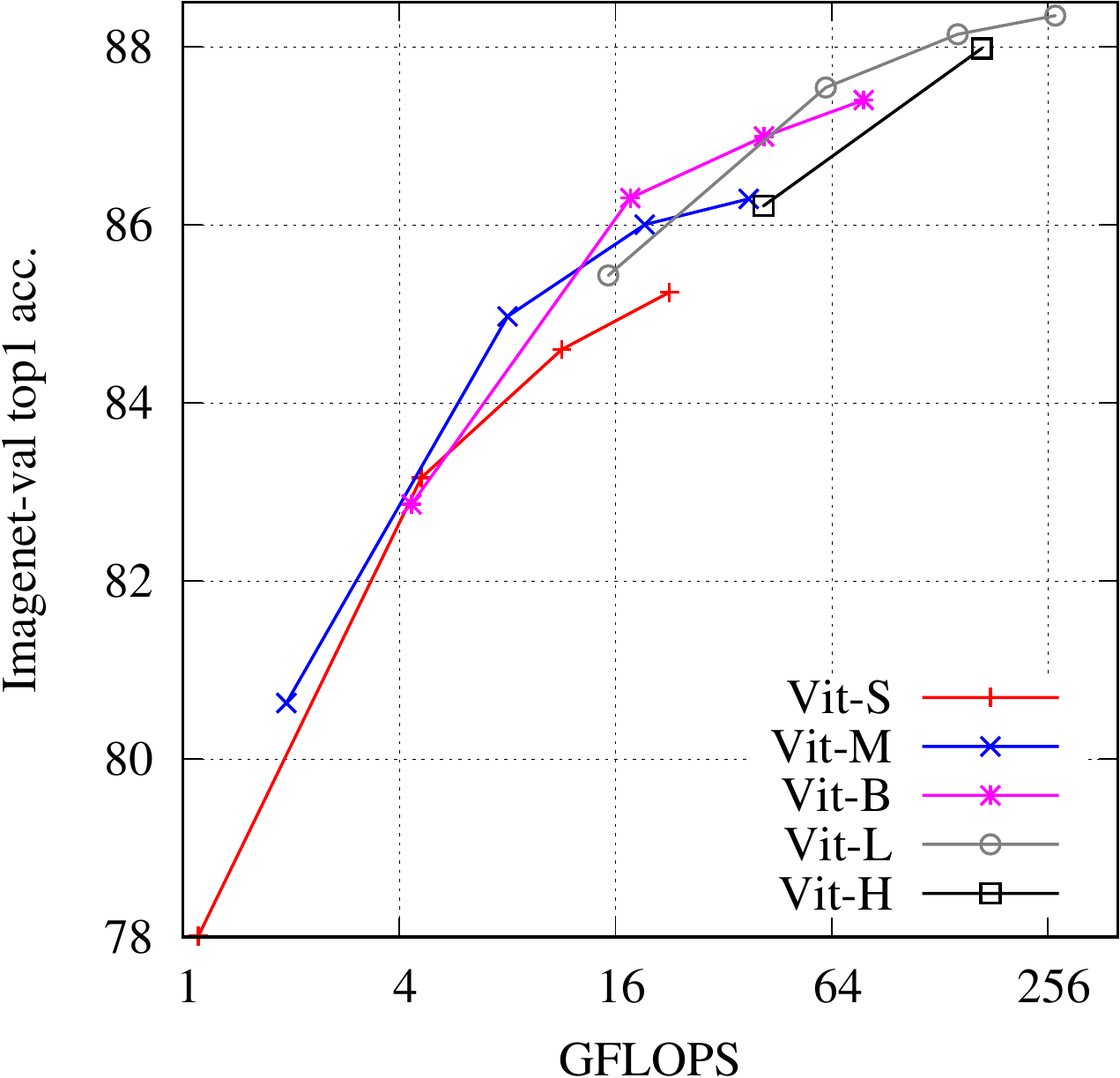}
    \hfill
    \includegraphics[height=0.25\linewidth,trim={5em 0 0 0},clip]{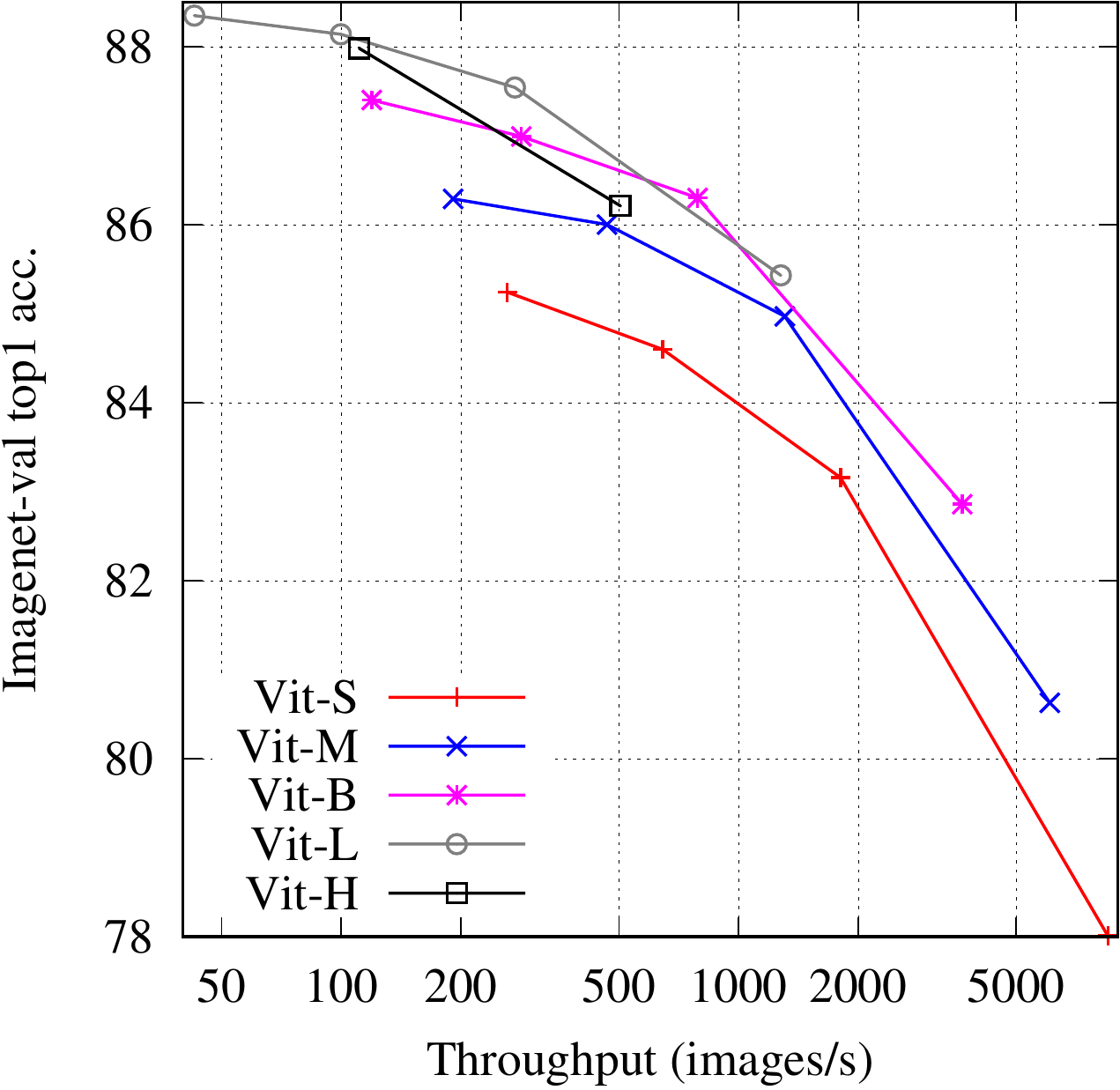}
    \hfill
    \includegraphics[height=0.25\linewidth,trim={5em 0 0 0},clip]{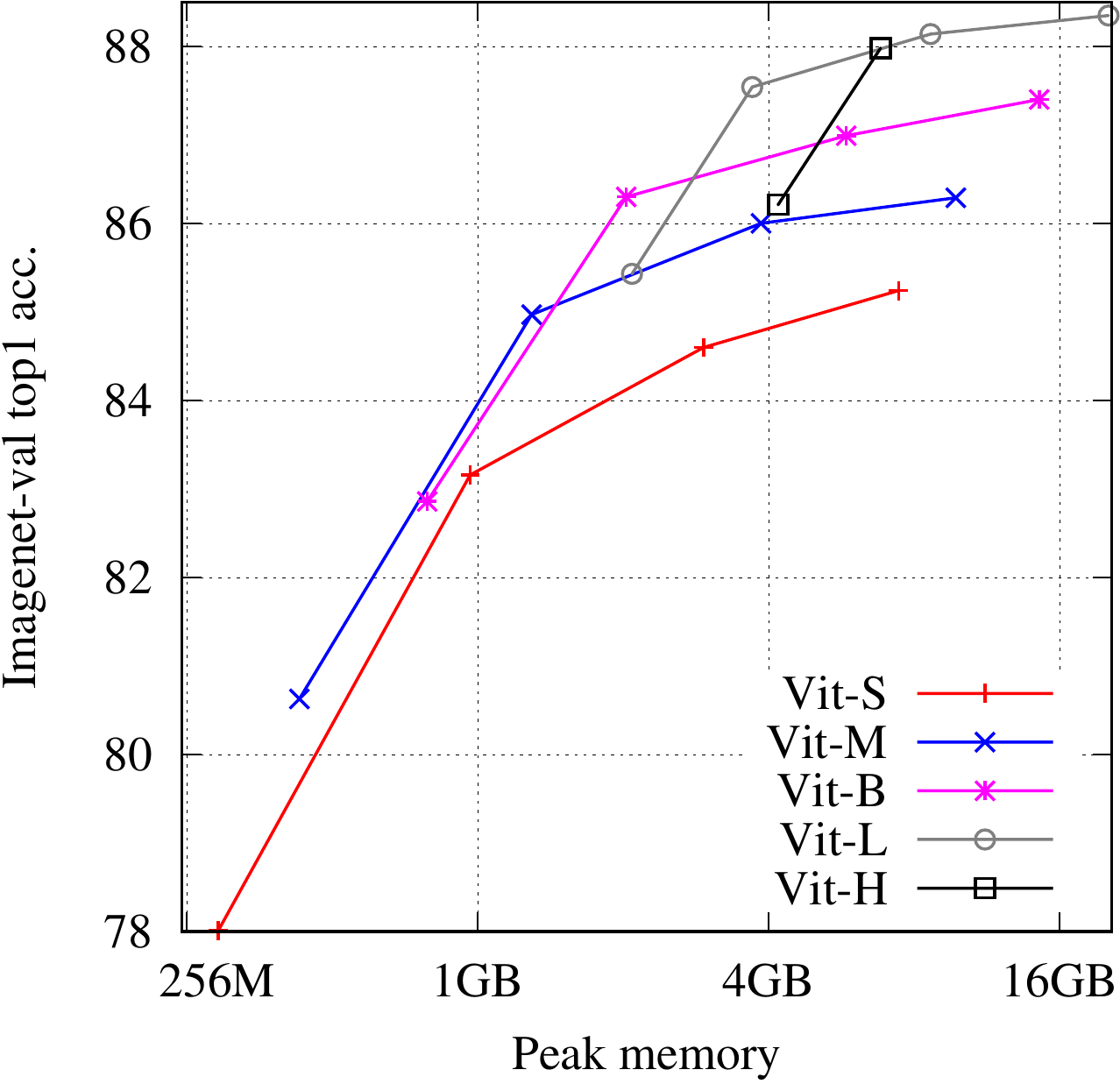}
    \end{minipage}
    \medskip
    
    \begin{minipage}[b]{1.0\linewidth}
    \includegraphics[height=0.25\linewidth]{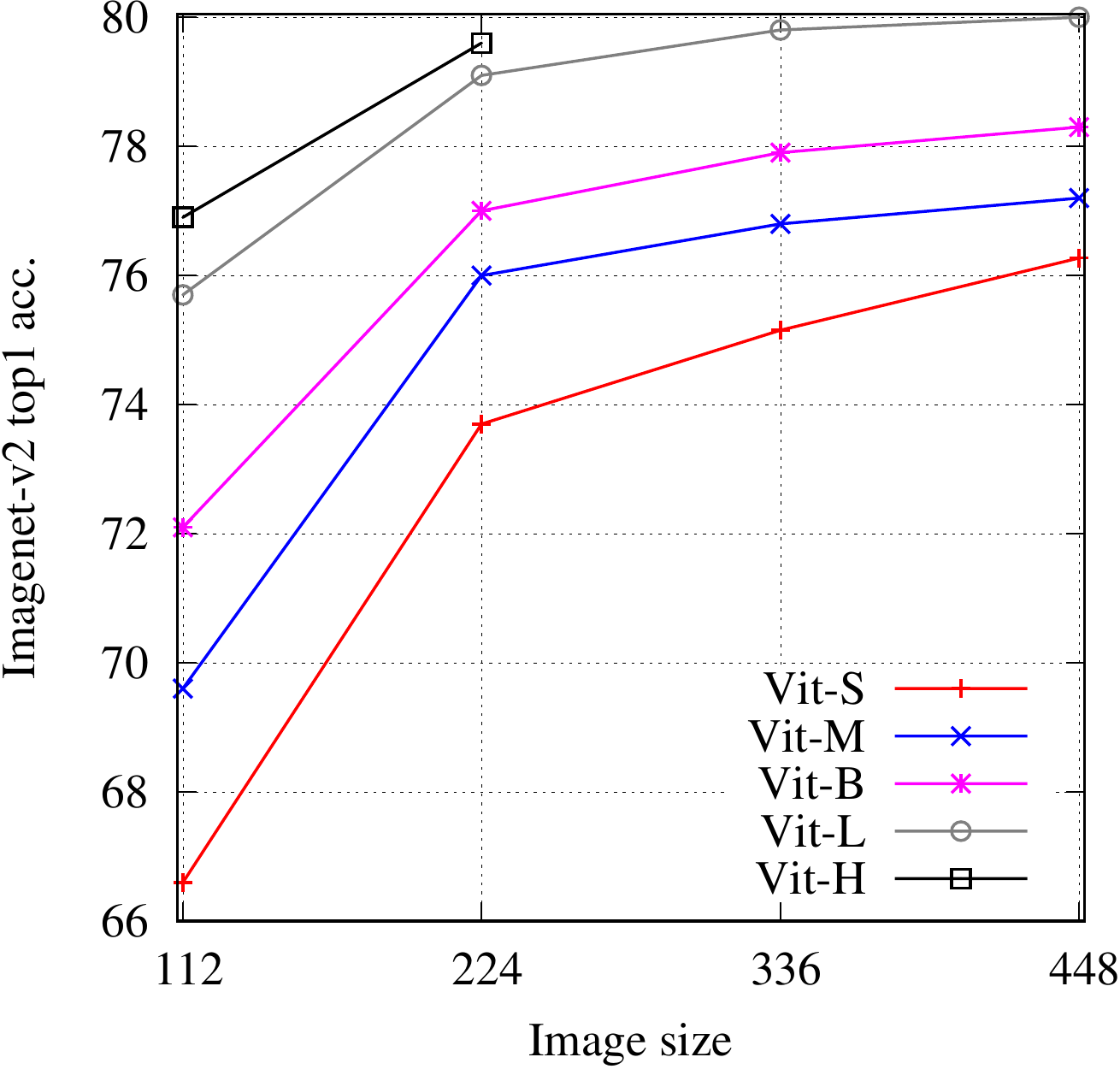}
    \hfill
    \includegraphics[height=0.25\linewidth,trim={5em 0 0 0},clip]{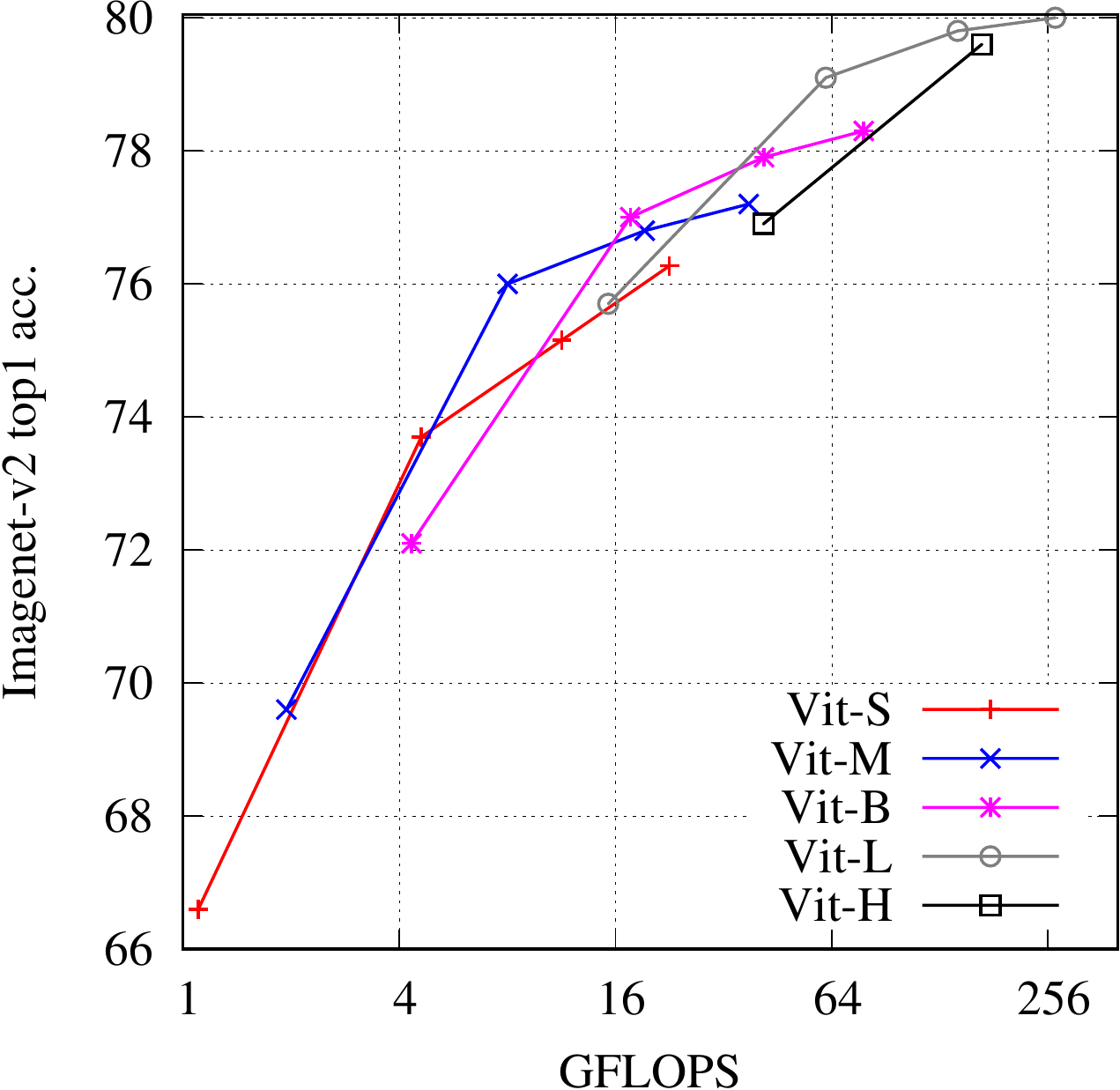}
    \hfill
    \includegraphics[height=0.25\linewidth,trim={5em 0 0 0},clip]{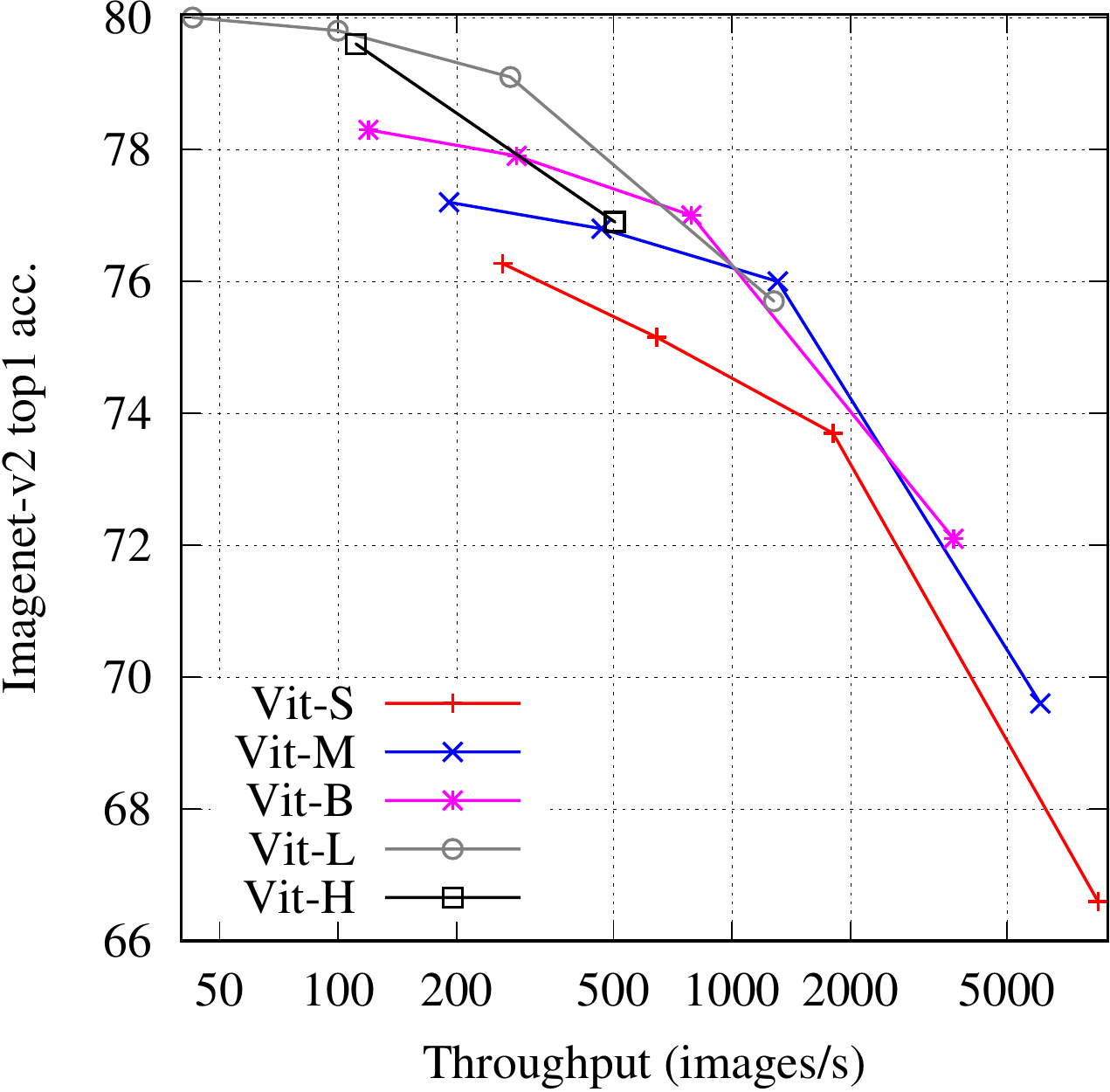}
    \hfill
    \includegraphics[height=0.25\linewidth,trim={5em 0 0 0},clip]{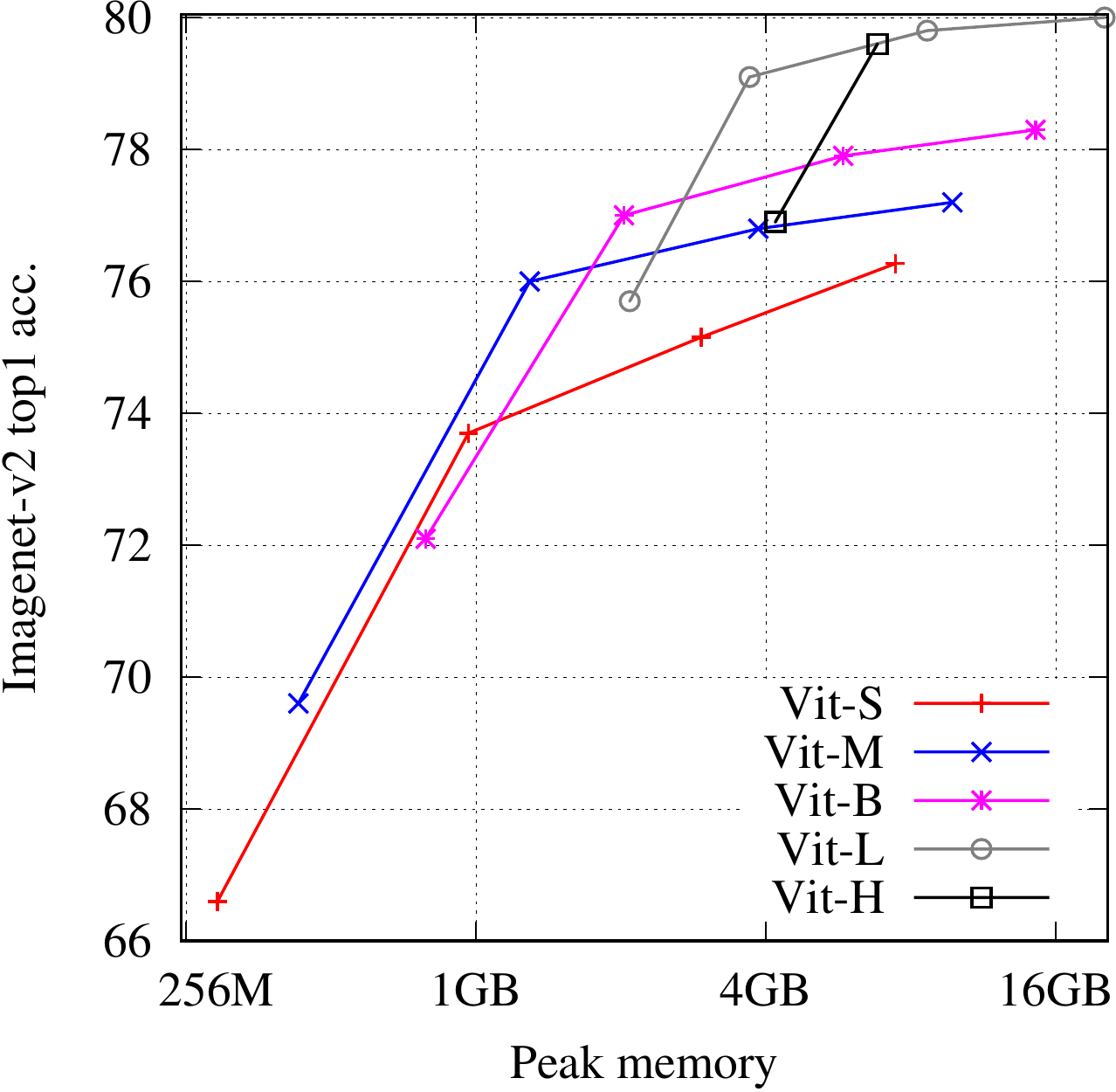}
    \end{minipage}
    
    \caption{\textbf{Flops/accuracy trade-offs:} We measure the accuracy on (\emph{top}) Imagenet1k-val and (\emph{bottom}) Imagenet-v2 as a function of different measures of complexity, which we vary for each model by increasing the resolution: 112\,$\times$\,112, 224\,$\times$\,224, 336\,$\times$\,336, 448\,$\times$\,448 (except for ViT-H, where we stop at 224\,$\times$\,224). All those models were pre-trained on Imagenet21k and fine-tuned on Imagenet1k. 
\label{fig:tradeoffs}}
\end{figure*}

Since both the model size and the resolution increase the accuracy and the complexity, the question is  which combination (model,resolution) we should  use. 
This question was noticeably analyzed by Bello \etal~\cite{Bello2021RevisitingRI} for ResNet, who pointed out that the Pareto-optimal resolution is typically lower than what was employed when the measure of complexity are FLOPS. 
We report trade-offs for different complexity measures in Figure~\ref{fig:tradeoffs}, see also Table~\ref{fig:resolution} in the main paper. 
Selecting a ViT operating at resolution 224 generally seems a good strategy. It is unclear whether this choice is absolutely good, or if it is better just because most of the hyper-parameter tuning effort in this paper and previous ones has been carried out at this specific resolution.

\subsection{Comparison with BerT-like approaches}
Although it is fully supervised, our  \ours method shares some similarities with purely self-supervised approaches such as  DINO~\cite{caron2021emerging}, MAE~\cite{He2021MaskedAA}, or BeiT~\cite{bao2021beit}.
Indeed, the \ours loss on submodels can be seen as an unsupervised or self-supervised loss.  
In contrast to \ours,  DINO does not backpropagate on the model that serves as a teacher, since this one is obtained by EMA: we cannot differentiate because it is based on past models that are not stored anymore. 

In Table~\ref{tab:comp_ssl} we compare \ours with BerT-like pre-training approaches as they are known to be very effective with vision transformers. We  find that our approach outperforms these competitive approaches when we can pre-train on Imagenet21k. 
Our \ours approach could potentially be used for unsupervised training, or to  finetune  self-supervised models: for instance, BeiT is typically fine-tuned on Imagenet-21k during a large number of epochs. 
We leave this exploration to future work.

\begin{table}
    \centering
   
    \scalebox{0.79}{
    \begin{tabular}{@{\ }c@{\ \ }cc@{\ \ }ccc|ccc@{\ }}
    \toprule
         & \multirow{2}{*}{Model} & 
        \multirow{2}{*}{Method}  &
        \#pretraining & \#finetune & \multicolumn{3}{c}{ImageNet}   \\
         & &  & epochs & epochs & val & Real & V2  \\
        \midrule
        \multirow{12}{*}{\rotatebox{90}{Imnet-1k pre-/training}} & \multirow{5}{*}{ViT-B}
         & \multirow{2}{*}{BeiT} & 300 & $100^{(1k)}$ & 82.9  & \_ & \_ \\
        & &  &    800 & $100^{\small(1k)}$ & 83.2  & \_ & \_\\
        \\[-0.8em]
        & & {\pzo}MAE$^\star$   & 1600 & $100^{(1k)}$ & 83.6  & 88.1 &  73.2  \\
                \\[-0.8em]
        & &  \cellcolor{blue!7}  & \cellcolor{blue!7} $400^{(1k)}$  & \cellcolor{blue!7} $20^{(1k)}$ & \cellcolor{blue!7}83.8  & \cellcolor{blue!7}88.6 & \cellcolor{blue!7}73.5  \\
        
        & &  \multirow{-2}{*}{\cellcolor{blue!7} Ours} & \cellcolor{blue!7} $800^{(1k)}$ & \cellcolor{blue!7} $20^{(1k)}$ & \cellcolor{blue!7}\textbf{84.2}  & \cellcolor{blue!7}\textbf{88.5} & \cellcolor{blue!7}\textbf{74.2}  \\
        
        \cmidrule{2-8}
        
        & \multirow{7}{*}{ViT-L}
        & BeiT   & 800 & $30^{(1k)}$ & 85.2  & \_ & \_  \\
                \\[-0.8em]
      &  & \multirow{3}{*}{MAE} & 400 & $50^{(1k)}$ & 84.3  & \_ & \_ \\
       &  &  & 800 & $50^{(1k)}$ & 84.9  & \_ & \_ \\
      &  &  & 1600 & $50^{(1k)}$ & 85.1  & \_ & \_ \\
              \\[-0.8em]
      & & {\pzo}MAE$^\star$   & 1600 & $50^{(1k)}$ & \textbf{85.9}  & \textbf{89.4} & \textbf{76.5} \\
              \\[-0.8em]
       & & \cellcolor{blue!7}   &  \cellcolor{blue!7} $400^{(1k)}$ & \cellcolor{blue!7} $20^{(1k)}$ & \cellcolor{blue!7}85.0  & \cellcolor{blue!7}89.4 &  \cellcolor{blue!7}75.5  \\  
       & & \multirow{-2}{*}{\cellcolor{blue!7} Ours}   & \cellcolor{blue!7} $800^{(1k)}$ & \cellcolor{blue!7} $20^{(1k)}$ & \cellcolor{blue!7}85.3   & \cellcolor{blue!7}89.2  &  \cellcolor{blue!7}75.5    \\  
     \midrule
     
     \multirow{8}{*}{\rotatebox{90}{Imnet-21k pre-training}} & \multirow{4}{*}{ViT-B}
      &  \multirow{2}{*}{BeiT}  &  150 & $50^{(1k)}$  & 83.7  & 88.2 & 73.1  \\
         &  &  &  150 + $90^{(21k)}$ &   $50^{(1k)}$ & 85.2  &  89.4 & 75.4  \\
        \\[-0.8em]
    & &  \cellcolor{blue!7} &  \cellcolor{blue!7} $90^{(21k)}$ &   \cellcolor{blue!7} $50^{(1k)}$ &  \cellcolor{blue!7}86.0 &  \cellcolor{blue!7} \textbf{89.8} &   \cellcolor{blue!7}\textbf{77.0}  \\
     & & \multirow{-2}{*}{\cellcolor{blue!7} Ours}   &  \cellcolor{blue!7} $270^{(21k)}$ &   \cellcolor{blue!7} $50^{k)}$ &  \cellcolor{blue!7}\textbf{86.3}  &  \cellcolor{blue!7}89.7 &   \cellcolor{blue!7}\textbf{77.0}  \\
    \cmidrule{2-8}
     & \multirow{3}{*}{ViT-L}
     & \multirow{2}{*}{BeiT} &  150 & $50^{k)}$  & 86.0  & 89.6 & 76.7  \\
    &  &  &  150 + $90^{1k)}$ &   $50^{k)}$ & \textbf{87.5}  &  90.1 & 78.8  \\
        \\[-0.8em]  &  & \multirow{-1}{*}{\cellcolor{blue!7} Ours}   &  \cellcolor{blue!7} $90^{1k)}$ &  \cellcolor{blue!7} $50^{k)}$ &   \cellcolor{blue!7}\textbf{87.5}  &  \cellcolor{blue!7}\textbf{90.3} &  \cellcolor{blue!7}\textbf{79.1} \\
     \bottomrule
    \end{tabular}}
    
    \vspace{-0.5em}
 \caption{Comparison of self-supervised pre-training with our supervised approach. %
 All models are evaluated at resolution $224\!\times\!224$. 
 We report image classification results on ImageNet val, real and v2 in order to evaluate overfitting. $^{(21k)}$ indicate a finetuning with labels on ImageNet-21k and $^{1k)}$ indicate a finetuning with labels on ImageNet-1k. $^\star$ indicates the improved setting of MAE using pixel (w/ norm) loss. 
MAE training is more efficient for a given number of epochs, thanks to its masking strategy. }
    \label{tab:comp_ssl}    
\end{table}

\section{Efficient stochastic depth} 
\label{app:esd}

\paragraph{Quantization of stochastic depth rate.}

\begin{figure}
    \centering
    \includegraphics[height=0.55\linewidth]{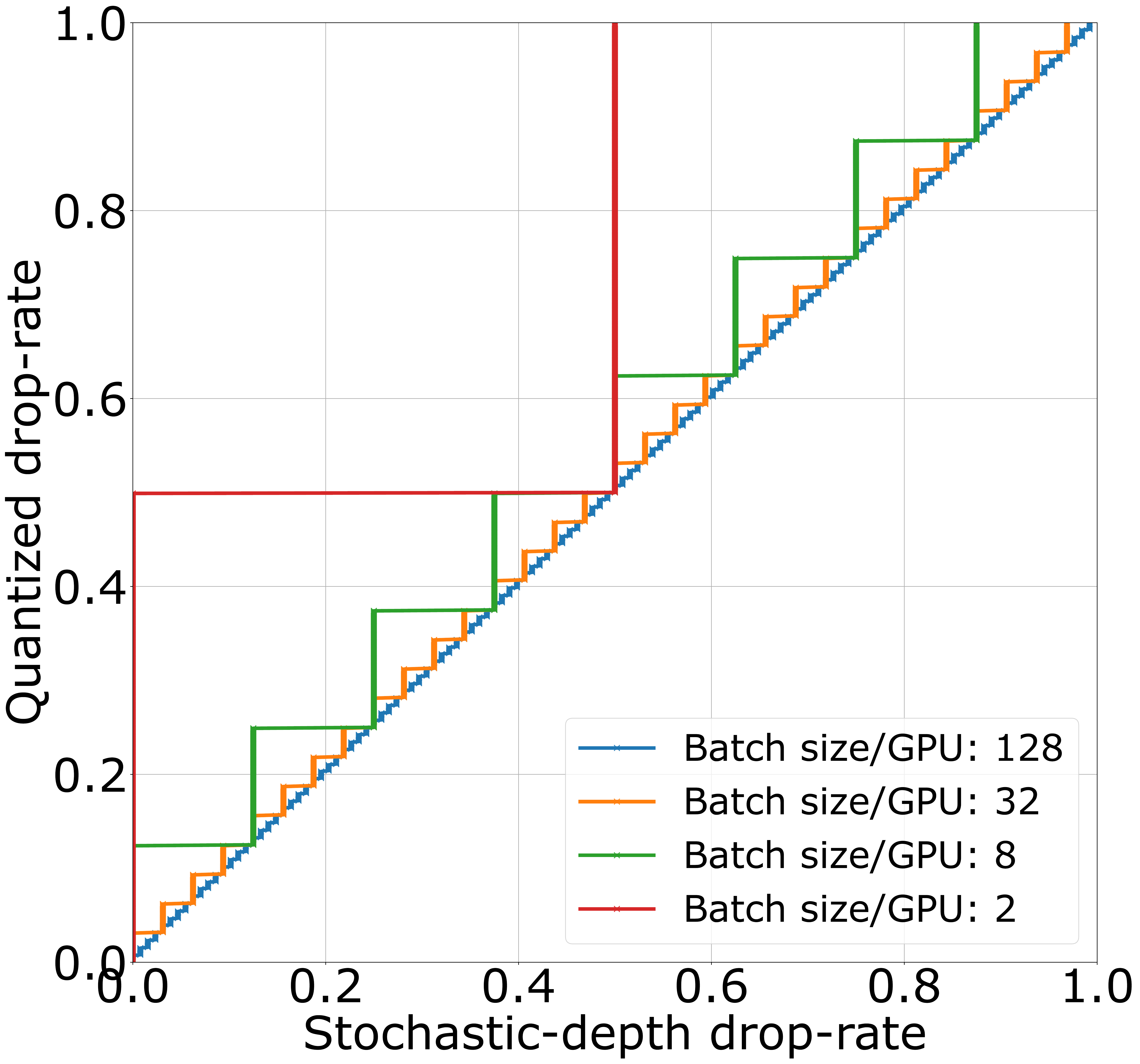}
    \caption{Efficient stochastic depth: we measure the effect of quantization on the effective drop path rate depending on the batch-size. 
    \label{fig:quant}}
        \vspace{-1em}
\end{figure}

Our ESD variant of stochastic depth determines a reduced batch size per GPU as follows: we multiply the input local batch size per GPU (LBS) by the requested drop-path hyper-parameter $\tau$, which produces the actual local batch size as
\begin{align}
    \mathrm{LBS}_\mathrm{ESD} = \lfloor \tau \times \mathrm{LBS} \rfloor. 
\end{align}
If LBS is large enough, this rounding has little effect. However, for large models or high-resolution images, the local batch size can become small, thereby leading to a more aggressive rounding when computing $\mathrm{LBS}_\mathrm{ESD}$. This leads to a coarse approximation of the stochastic depth parameter $\tau$, as shown in Figure~\ref{fig:quant}. 
For example, for a local GPU batch size of 8, the only possible values of $\tau$  are: $\tau \in \{0, 0.125, 0.250, 0.375, 0.5, 0.625, 0.75, 0.875, 1\}$, because the actual batch size is floored to an integer. In this figure, we report the mapping between the effective drop-rate and the actual one. 
In practice, this quantization effect must be taken into consideration in extreme cases (very large models or higher resolution images). In such cases, we compute the effective stochastic depth when computing the ratio $1/(1-\tau)$ for the inference-time model.

\section{Submodel analysis}
\label{sec:submodels}

\paragraph{Number of layers with stochastic depth. } 

\begin{figure}
    \centering
    \includegraphics[height=0.65\linewidth]{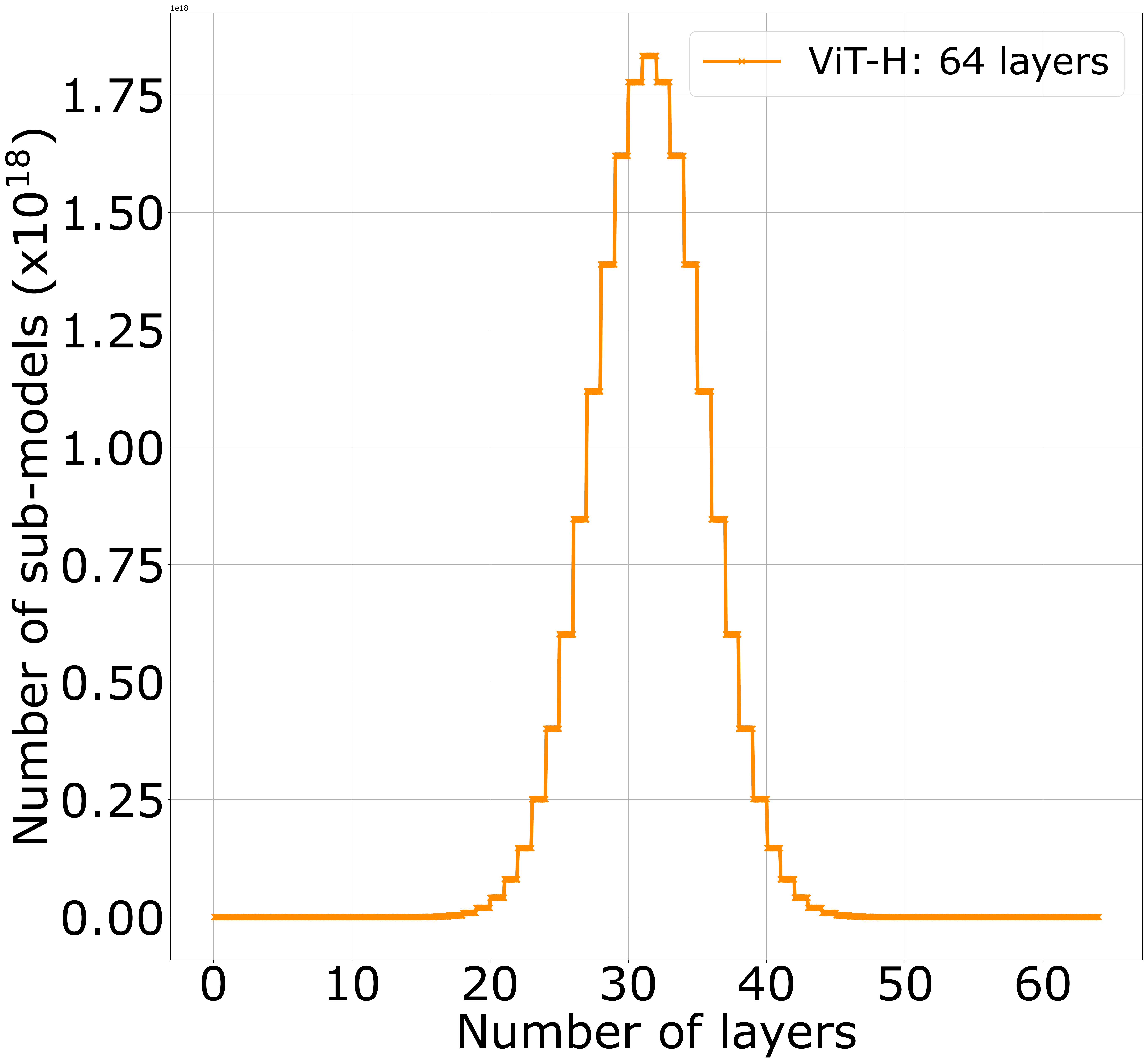}
    \caption{Number of submodels with a given number of layers for ViT-H (32 blocks). $\tau$\,=\,0.5 gives the same probability to instantiate each. 
    \label{fig:nb_layers_kept}}
\end{figure}

In Figure~\ref{fig:nb_layers_kept}, we show the number of submodels exist for a given number of layers. This corresponds to a binomial distribution, which attains its maximum with 32 layers. 
Setting $\tau$\,=\,0.5 gives the same probability to instantiate each of those, hence we can se that it stochastic depth will draw with very high probability a model that contains about 32 layers. 

\paragraph{Layer ablation.} 

\begin{figure}
    \begin{minipage}[b]{1.0\linewidth}
    \includegraphics[width=\linewidth,trim={0 0.5cm 0 0},clip]{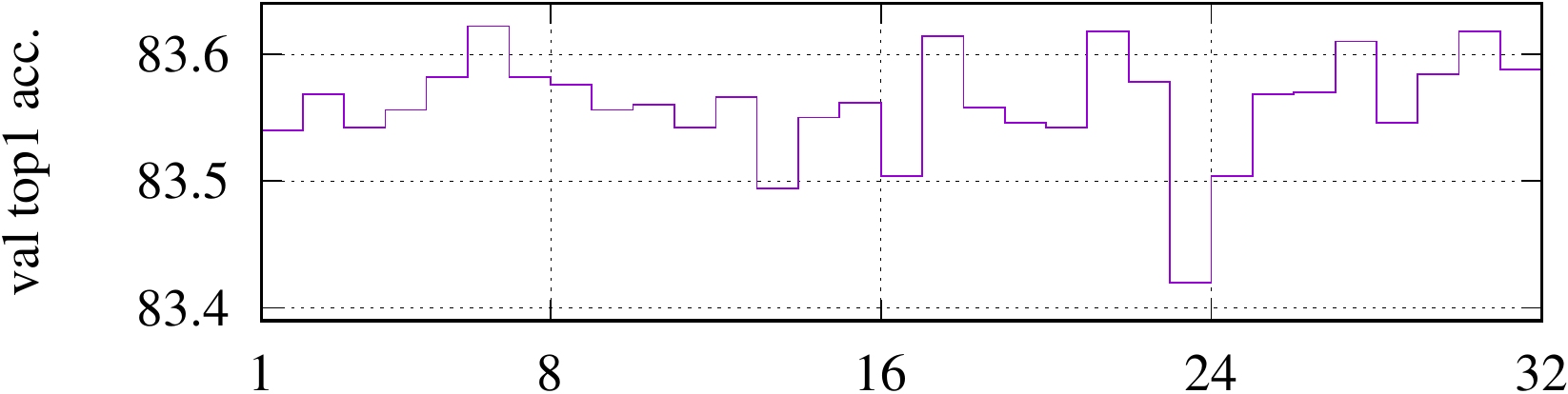} \\
    \includegraphics[width=\linewidth]{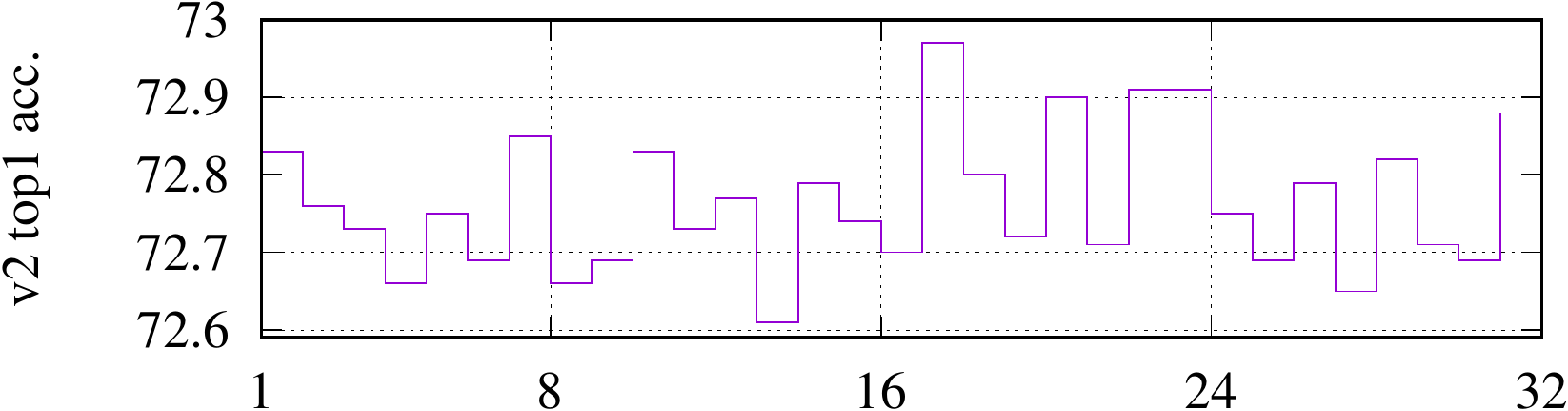} 
    \end{minipage}
    \caption{Layer ablation: we trim one block (i.e., a multi-head self-attention and its corresponding ``FFN'') of a fixed ViT-H network 126\,$\times$\,126 learned with submodel co-training, and evaluate the performance of the $L$ corresponding subnetworks. Variations are overall small and there is no strong correlation between Imagenet-val and -v2 accuracy.  %
\label{fig:layerablation}}
\end{figure}

Inspired  by experiments from Fan et al.~\cite{fan2019reducing}, we measure the performance of the submodels produced when dropping exactly one block (Multi-head soft-attention and corresponding Feedforward) from a trained ViT model, in this case a ViT-H trained at resolution 126\,$\times$\,126 on Imagenet1k. Our objective is to measure whether some layers are more important than others. The performance of the submodels are reported in Figure~\ref{fig:layerablation}. We observe that almost most of the submodels have almost an identical performance on Imagenet1k-val, around 83.55\% in top-1 accuracy, close to the performance (83.6\%) of the full 32-blocks/64-layers model. 

\end{document}